\documentclass[10pt,twocolumn,letterpaper]{article}

\usepackage[pagenumbers]{cvpr} %

\definecolor{cvprblue}{rgb}{0.21,0.49,0.74}
\usepackage[pagebackref,breaklinks,colorlinks,allcolors=cvprblue]{hyperref}
\usepackage{wrapfig}
\usepackage{graphicx}
\usepackage{xcolor}
\usepackage{caption}
\usepackage{subcaption}
\usepackage{hyperref}
\usepackage{url}
\usepackage{enumitem}
\usepackage{amssymb}

\def\paperID{6088} %
\def\confName{CVPR}
\def\confYear{2026}

\title{Toward A Better Understanding of Monocular Depth Evaluation}

\author{Siyang Wu, Jack Nugent, Willow Yang, Jia Deng\\Princeton University}

\begin{document}
\maketitle

\begin{abstract}
Monocular depth estimation is an important task with rapid progress, but how to evaluate it is not fully resolved, as evidenced by a lack of standardization in existing literature and a large selection of evaluation metrics whose trade-offs and behaviors are not fully understood. This paper contributes a novel, quantitative analysis of existing metrics in terms of their sensitivity to various types of perturbations of ground truth, emphasizing comparison to human judgment. Our analysis reveals that existing metrics are severely under-sensitive to curvature perturbation such as making smooth surfaces bumpy. To remedy this, we introduce a new metric based on relative surface normals, along with new depth visualization tools and a principled method to create composite metrics with better human alignment. Code and data are available at: \href{https://github.com/princeton-vl/evalmde}{https://github.com/princeton-vl/evalmde}.
\end{abstract}

\vspace{-2mm}
\section{Introduction}
\vspace{-1mm}
Monocular depth estimation (MDE) is the task of estimating pixelwise depth from a single RGB image. It has become a standard task with rapid progress, due to its importance for many applications, such as robotics, AR/VR, and the generation of images, videos or 3D worlds. 

However, how to evaluate monocular depth estimation is not fully resolved. This is reflected by a lack of standardization in existing literature, and a very large menu of evaluation metrics. It is in fact possible for two papers (e.g., \citep{bochkovskiydepthpro,wang2025moge}) to report 5 to 7 different metrics each but with zero overlap\footnote{Two metrics can share the same name but are different due to differences in depth alignment.}.

Even when models are compared under the same set of metrics, it is often not clear how to interpret the results. 
If a model performs well under one metric (say AbsRel after affine alignment on disparity) but poorly under another (say $\delta^1$ with no alignment), what does it mean? Often there lack clear answers, because the trade-offs between different metrics are not fully understood. 

The large selection of metrics arises from a combination of decisions that need to be made when designing an evaluation metric. For example, the decisions can include how to compute error of individual depth values against ground truth (by difference in depth, log depth, or inverse depth), whether and how to discount error for faraway points, whether to binarize the error and if so with what threshold, and how to account for the unknown global scale through alignment (in depth or inverse depth), and whether to also allow an unknown global shift. %

The complexity of these decisions is due to several issues inherent to monocular depth estimation that preclude straightforward comparisons of individual depth values. One is scale ambiguity. The global scale of a scene is fundamentally ambiguous: a scene can be a miniature replica of a larger identical scene. This scale ambiguity can also occur locally in a scene: it can be impossible to tell from a single image whether a drone in the sky is a big one far away or small one close by. Even without any scale ambiguity, another issue is the unbounded range of possible depth values in the same scene---consider an ocean that extends to the horizon. In such cases, some form of normalization or weighting is necessary to prevent errors on faraway objects from dominating. \looseness=-1

To make matters worse, the predicted depth values alone do not reconstruct a 3D shape because it does not tell us the X and Y coordinates in 3D; for that we also need the camera intrinsics, which are often unknown and need to be predicted. But camera intrinsics can be ambiguous from a single image and the prediction can be reasonable but off. And the predicted depth and predicted camera intrinsics can together give a reasonable 3D shape but have large errors when evaluated separately. \looseness=-1

The discussion above suggests that the evaluation of monocular depth is inherently complex. The existence of many metrics reflects best efforts to address such complexity. On the other hand, there has been limited study on the trade-offs of these metrics, and research on monocular depth could benefit from additional investigation.  

In this paper we seek to improve our understanding of monocular depth evaluation. Our goal is to study various evaluation metrics and shed light on their trade-offs and behaviors, and to develop a principled method to customize or combine metrics to align with particular preferences, such as those based on human perception. 

It is important to note that our goal is \emph{not} a new ``perfect'' metric that is better in all aspects and supersedes all existing metrics. A single ideal metric probably does not exist and the best metric is likely application dependent. We only seek to offer additional insights on the behaviors of existing metrics and how they might be customized and combined toward particular goals.  

\noindent \textbf{Our Approach:} Our main approach is to quantify the sensitivity of each metric to various types of perturbations of the ground truth depth.  In particular, we focus on 6 types of perturbations---surface orientation, camera intrinsics, relative scale, curvature, affine transform, and boundary---which are interpretable and representative changes to ground truth depth that could shed light on the behavior of a metric. The definition of each perturbation is given in Section~\ref{sec:sensitivity_analysis}. 

Given these perturbations, we can measure the sensitivity of a metric to each perturbation, and compare the sensitivity between metrics. The basic idea is to measure the numerical responses of a metric under various perturbations, and compare such responses across metrics, after proper normalization to account for the arbitrary choice of measuring units.

\noindent
\textbf{Human Sensitivity:} We also apply our sensitivity analysis to human judgment. Human judgment can serve as a potentially useful reference for sensitivity comparisons because (1) many generative applications produce content for human consumption and (2) the human visual system remains arguably the best general-purpose depth perception system and human judgment may provide clues on what is and is not important for achieving human-level visual capabilities. Note that we do \emph{not} claim that human judgment is the gold standard or an ideal proxy for all or some downstream applications; it is simply a potentially useful reference.  

We measure the sensitivity of human judgment to various perturbations. Using a collection of synthetic test scenes and varying amounts of perturbations, we ask human annotators to judge whether a (possibly) perturbed depth map is the ground truth of a given RGB image. We then average the binary annotations to estimate the exchange rates of human judgment versus other metrics. 

To help human annotators examine the reconstructed geometry, we introduce two new visualization tools: Textureless Relighting and Projected Contours. These new tools overcome the limitations of conventional visualizations such as textured point clouds, which can mask geometric artifacts because, like in video games, texture maps can create fake perceived geometry. Beyond measuring human sensitivity, the tools can also be useful in developing future depth models.  

Using human judgment as a reference yields interesting findings about existing metrics. Humans are sensitive to affine transforms of depth or disparity, but many metrics perform affine alignment and are thus completely insensitive.  %
Most notably, all widely used metrics have very poor sensitivity to curvature perturbation (e.g., making a smooth surface bumpy). 

\noindent
\textbf{Sensitivity Aligned Composition:} In addition to improving our understanding, our sensitivity analysis enables us to combine existing metrics to align with a specific sensitivity profile, such as that of humans or a downstream application. The basic idea, which we call ``sensitivity aligned composition (SAC)'', is to combine existing metrics through some parametric form of composition such as a weighted average, and optimize the composition parameters such that the combined metric achieves the desired sensitivities to a given set of perturbations.  

Because existing metrics are overly insensitive to curvature perturbation compared to human judgment, we propose a new metric RelNormal, which is based on relative surface normals and thus sensitive to curvature perturbation. Combining RelNormal and a subset of existing metrics, we propose SAWA-H (Sensitivity Aligned Weighted Average based on Human judgment), a new, composite metric that aligns better with human judgment than all existing metrics. SAWA-H can supplement existing metrics in cases where alignment with human judgment is desired. 

\noindent
\textbf{Summary of Contributions: } 
Our contributions are three fold. (1) We conduct a novel, quantitative sensitivity analysis of the commonly used metrics under various perturbations. (2) We introduce two new tools of depth visualizations and measure sensitivity of human judgment, revealing that existing metrics are overly insensitive to curvature perturbation. (3) We introduce SAWA-H, a new metric that better aligns with human judgment via optimized combination of a set of base metrics; we also introduce RelNormal, a new base metric designed to enable better human alignment. Code and data are available at: \href{https://github.com/princeton-vl/evalmde}{https://github.com/princeton-vl/evalmde}.

\section{Related Work}
\textbf{Monocular Depth Evaluation Metrics.} Conventional metrics directly evaluate the difference between predicted and ground truth depth. They evaluate accuracy of ordinal relationship \citep{zoran2015ordinalrelationship,chen2016diw} or per-pixel depth difference \citep{eigen2014depth,saxena2008make3d}. Recently, \citet{koch2018evaluation,chen2019oversmooth,ornek20222dto3d,talker2024mindedgerefiningdepth,pham2024sharpdepthsharpeningmetricdepth} noted that these metrics are not sensitive to over-smooth boundaries and error at predicting planes, and new metrics were proposed to evaluate boundary sharpness \citep{koch2018evaluation,chen2019oversmooth,bochkovskiydepthpro} and difference in 3D \citep{koch2018evaluation,ornek20222dto3d,wang2025moge}. Though prior work noted that existing metrics are not sensitive to some errors, there lacks a principled way to systematically study behavior of different metrics. We attempt to fill this gap. %

\noindent
\textbf{Monocular Depth Estimation Methods.} There is rapid development in MDE in recent years. It is worth noting that a decent number of methods predict affine-invariant depth or disparity \citep{birkl2023midas,yang2024depthanything,yang2024depthanythingv2,ke2024marigold,fu2024geowizard}, and during evaluation, they perform affine alignment on depth or disparity before computing metrics. \citet{li2025benchdepth} pointed out two issues with alignment: unfair comparison between methods that use different alignment, and alignment is sensitive to outliers. In our analysis we further find that affine-aligned metrics are misaligned with human judgment, which is sensitive to geometric distortions caused by affine perturbations of depth. \looseness-1

\noindent
\textbf{Human Sensitivity and Visualization Tools} Though human vision is known to be accurate and fairly robust, humans still exhibit peculiarities when judging 3D structure. When constructing the OASIS dataset, \citet{chen2020oasis} observed that annotators judged the shape correctly (i.e. the relative normals) but often made mistakes when estimating the overall orientation. \citet{linsley2025the} find that though humans achieve similar scores as deep neural networks  when estimating depth order, humans are much better at visual perspective taking (answering ``can object A see object B?''). This suggests that standard 3D geometric evaluations fail to capture the aspects of a scene that humans are sensitive to.

When visualizing depth, monocular depth estimation works typically plot heatmaps of either depth or disparity (\cite{birkl2023midas,eigen2014depth,zoran2015ordinalrelationship}). Other works also display novel views of the unprojected depth map with the points colored by the corresponding pixel in the original image \cite{wang2025moge,yin2021learning}. Notably, the shading in these point clouds does not correspond with the true geometry. \citet{wang2025moge}, \citet{BiNi}, and others additionally display a gray mesh object with shadows and specular highlights rendered from the original view. This is significant, as \citet{liu2004perceptual} finds that shadows, specular highlights, and other aspects of normal shading greatly improve humans' ability to discern curvature. We further expand on these visualization tools in section \ref{sec:visualization_tool}.

\section{Sensitivity Analysis}\label{sec:sensitivity_analysis}

\noindent\textbf{Definition of evaluation metric} Without loss of generality, we define a evaluation metric to be a function that maps a question and an answer to a non-negative score that represents the quality of the answer. For monocular depth, the question is a RGB image and the answer is a depth map. Sometimes the answer could also include predicted camera intrinsics. 

This definition is general and does not assume access to a ground truth answer, because there may not be a unique correct answer. The metric can be a neural network or a human. On the other hand, an evaluation metric, like all existing monocular depth metrics, can use privileged information including a ground truth answer and return a score by comparing the answer to the ground truth. 

\noindent\textbf{Inherent Ambiguity in Monocular Depth:}
It is worth noting that monocular depth estimation has inherent ambiguity: there always exist multiple possible 3D shapes that project to the same 2D image, and some are considered more plausible only because of priors. For example, the input image can depict an actual castle or a printed photo of the same castle, and there can be no way to tell them apart. 

In addition, the actual physical scene that generated the 2D image can be unreasonable as ground truth. Consider the Ames room~\cite{wikipedia:ames_room}, which is trapezoidal-shaped but appears rectangular to the human eye. It would be unreasonable for the ground truth to be trapezoidal, because this would lead to a monocular depth system that reconstructs trapezoids even for actual rectangular rooms. 

This inherent ambiguity suggests that (1) it is more fitting to use the more general definition of evaluation metric that does not assume a single known ground truth, and (2) human perception can be a reasonable reference ``ground truth'', because the inherent ambiguity makes it impossible to always correctly recover the actual physical ground truth. 

\begin{figure*}[htbp]
    \centering
    \hspace{0.05\textwidth}
    \begin{minipage}{0.30\textwidth}
        \centering
        \includegraphics[width=\textwidth]{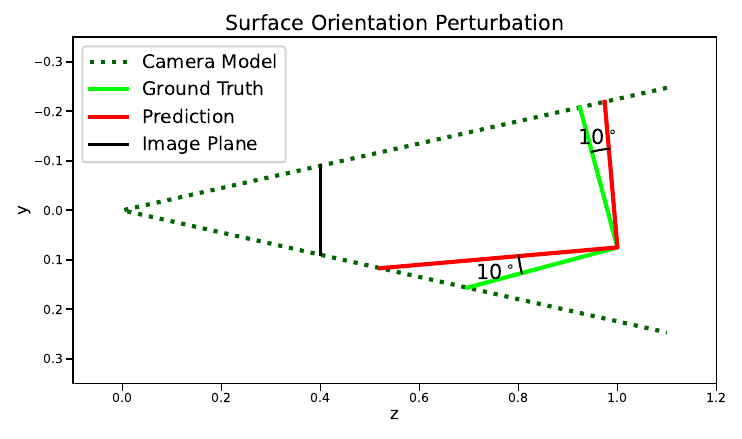}
    \end{minipage}
    \hfill
    \begin{minipage}{0.52\textwidth}
        \centering
        \includegraphics[width=\textwidth]{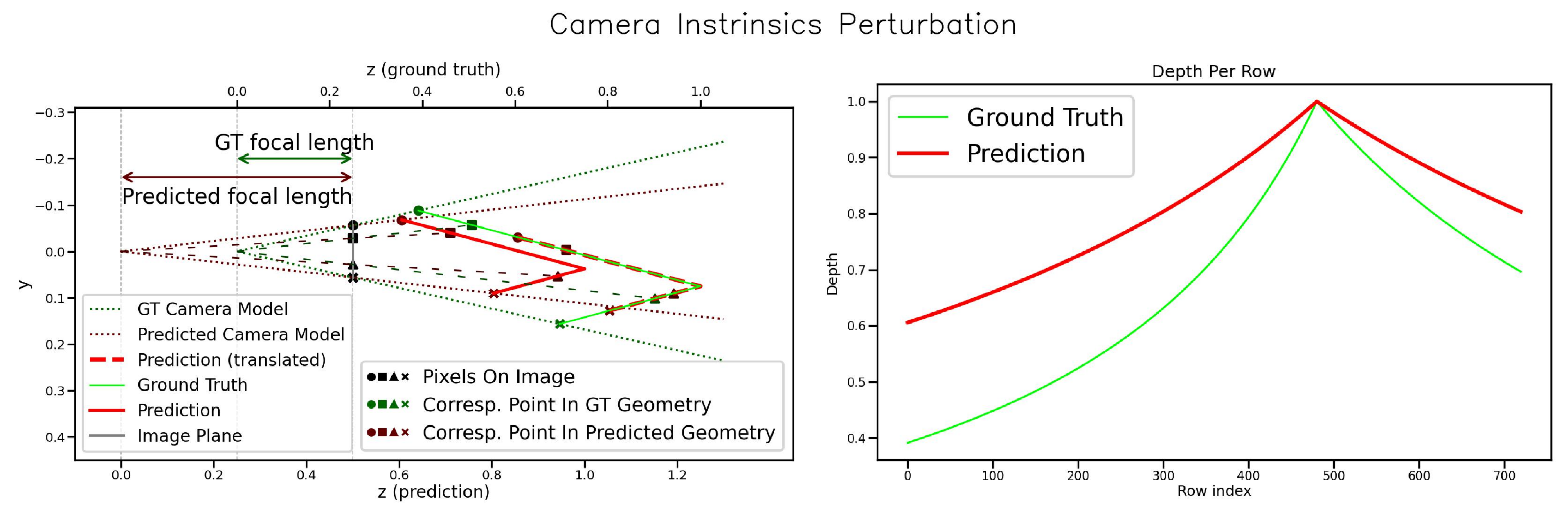}
    \end{minipage}
    \hspace{0.06\textwidth}
    \caption{\textbf{Left:} Surface orientation perturbation (side view). Orientation of prediction (red) equals rotating that of ground truth (green) around x-axis by $-10^\circ$. In side views through the paper, we show simple geometry whose depth is the same on each row. \textbf{Middle \& Right:} Camera intrinsics perturbation. Middle (side view of 3D shape): Prediction is made under wrong focal length (2x ground truth focal length), while predicted geometry (red) has shape similar to ground truth (green). Predicted geometry can overlap well with ground truth through translation (red dashed). Right (depth value of each row): Predicted depth substantially differs from ground truth depth. \looseness=-1}
    \label{fig:illustration_of_error_types}
    \vspace{-2mm}
\end{figure*}

\noindent
\textbf{Perturbations:} We study the sensitivity of metrics to a set of interpretable perturbations.
\begin{itemize}[leftmargin=1em, itemsep=1pt]
\item \textbf{Surface orientation perturbation} refers to perturbing a depth map such that the surface normal of each pixel is rotated by the same amount (or close to the same amount). The motivation is that the absolute orientation of surfaces can be ambiguous to humans~\cite{chen2020oasis}. Fig. \ref{fig:illustration_of_error_types} Left shows an example. %
\item \textbf{Camera intrinsics perturbation} refers to perturbing both focal length and depth while keeping the 3D shape similar in terms of surface normals.  The motivation is that prediction of focal length can be off but the reconstructed shape can still be reasonable. Fig. \ref{fig:illustration_of_error_types} Middle \& Right shows an example: although the 3D geometry is similar, the absolute difference between predicted depth and ground truth depth is large (Fig. ~\ref{fig:illustration_of_error_types} Right). %
\item \textbf{Relative scale perturbation} refers to perturbing relative scale between objects, e.g. scaling down the depth of a floating foreground object while keeping the background the same. The motivation is that the scale of a (floating) foreground object can be ambiguous to determine relative to the background. 
\item \textbf{Curvature perturbation} refers to perturbing the curvature of back-projected 3D surfaces, e.g. making a smooth surface bumpy. We perform curvature perturbation under two spatial frequencies. The motivation is the existence of smooth surfaces in nature (e.g., sand dunes, egg shells) and built environments. Please refer to the Appendix for details.\looseness=-1
\item \textbf{Affine transform perturbation} refers to perturbing ground truth depth by an affine transform of depth or disparity. The motivation is that many existing methods predict affine invariant depth. 
\item \textbf{Boundary perturbation} refers to blurring depth values across occlusion boundaries, resulting in distortion of geometry near boundaries. The motivation that some depth methods produce sharper boundaries, which are considered more desirable. 
\end{itemize}
Please refer to Appendix \ref{sec:perturbation_algo} for details of the perturbation algorithms and examples of perturbed depth.

\noindent
\textbf{Exchange rates:}  
For each metric and perturbation type, we measure the numerical response of the metric when applying that type of perturbation to ground truth. Then, we compare the responses of different metrics across different perturbation types to examine which metric is more or less sensitive to which perturbation. To facilitate this comparison, we introduce the notion of exchange rate.

The exchange rate between metrics A and B under perturbation P, $R(A;B|P)$ is (informally) defined as the ratio of changes of metrics A and B when perturbing the ground truth depth by the same intensity of perturbation P. 
Without loss of generality, we assume all metrics can be standardized such that the ground truth depth is given a score of 0 and the score increases as the depth deviates from ground truth.

For example, if after the same perturbation, the value of metric A (say RMSE) increases from 0 to 2 and value of metric B (say AbsRel) increases from 0 to 0.5, the exchange rate between metrics A and B is $2/0.5=4$. This rate can be interpreted as: to represent the same intensity of perturbation P, 4 units of metric A are needed, while we only need 1 unit of metric B. %

This intuitive notion of exchange rate is well defined if both metric A and B respond linearly to the intensity of the perturbation P. However, this is almost never the case. For example, some metrics are bounded and will eventually saturate for large perturbations. Therefore our formal definition of the exchange rate needs to also handle nonlinear functions.

Formally, we treat metrics $A(x)$ and $B(x)$ as smooth functions of intensity $x$ of perturbation P, and define the exchange rate $R(A;B|P)$ to be the ratio between their derivatives at zero: 
\begin{equation}
R(A;B|P) \triangleq A'(0)/B'(0).
\end{equation}

In other words, we approximate each metric as a linear function of perturbation intensity in a small neighborhood around zero. This approximation is justifiable because as models improve, the neighborhood around zero becomes more important.

In practice, we approximate each metric with the least squares fit of a quadratic function $ax^2 + bx$, and we take the derivative of the quadratic function at zero. Compared with a linear form, the quadratic form better approximates plateauing metrics and removes the need to manually select the size of the neighborhood around zero. 

\noindent
\textbf{Comparing Sensitivity:} It may appear that metric A is more sensitive than metric B to perturbation P if the exchange rate $R(A;B|P)$ is large, as metric A has a larger response. But this impression is false because units of metrics A and B can be arbitrarily chosen. For example, metric B and metric A can be identical except that A is measured in meters and B is in millimeters. Then the exchange rate is 1000,  but metrics A and B are effectively identical.  

This means that a single exchange rate $R(A;B|P)$ between A and B under perturbation P is not meaningful in isolation and must be interpreted in comparison with another exchange rate. Specifically, let $R(A;B|Q)$ be the exchange rate of the same metrics A and B under a new perturbation Q, using the same units of metric A and B. If the exchange rate under $P$ (for example 1000:1) is bigger than that under $Q$ (for example 2:1), then we can conclude that relative to B, metric A is more sensitive to perturbation P than Q. This is because 1 unit worth of perturbation P measured under metric B translates to 1000 units under metric A, whereas 1 unit worth of perturbation Q measured under metric B translates to only 2 units under metric A. In other words, compared to metric B, perturbation P causes larger numerical changes to metric A than perturbation Q.  

\noindent
\textbf{Dataset for measuring sensitivity.} To estimate the exchange rate in practice, we need a dataset of 3D scenes with ground truth depth maps. This is because the sensitivity of a metric is often scene dependent, in which case we obtain the expected sensitivity averaged over multiple scenes.  

We create a dataset of synthetic scenes using Infinigen~\cite{raistrick2023infinite,raistrick2024infinigenindoor}, a procedural generator of photorealistic nature and indoor scenes. We choose to use synthetic data because real-world data are limited in the diversity of scenes and availability of dense depth ground truth. For example, no pixelwise real-world ground truth is available for large natural scenes. We choose Infinigen because it covers both indoor and natural scenes and is easy to customize. 

The dataset consists of 95 scenes. Among them, 56 are indoor scenes, which have short depth range and more regular shape, and 39 are nature scenes, which have long depth range and less regular shape. For each perturbation type, we perturb by at least 6 different intensities, with a total of 5320 perturbed depth in the dataset. Example scenes are included in Appendix \ref{sec:dataset_example}.

\noindent
\textbf{An Example of Comparison of Existing metrics}
Here we show an example of using our methodology to compare sensitivity between 3 metrics: AbsRel-No Align (AbsRel \citep{saxena2008make3d} with no alignment), AbsRel-Disparity Af (AbsRel after affine alignment of disparity), and Boundary F1-No Align (Boundary F1 \citep{bochkovskiydepthpro} with no alignment).

\begin{figure}[t]
    \centering
    \includegraphics[width=\linewidth]{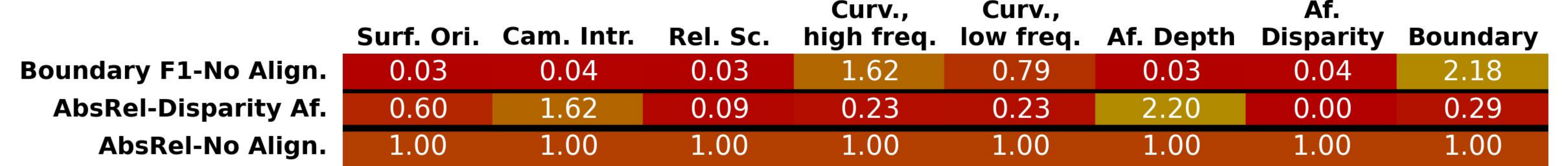}
    \caption{Exchange rate between each metric and AbsRel with no alignment. Because absolute scale of the exchange rate is not important, we normalize each row for better visualization. Perturbations from left to right: surface orientation, camera intrinsics, relative scale, curvature (high frequency), curvature (low frequency), affine transform of depth, affine transform of disparity, and boundary.}
    \label{fig:example_sensitivity_comparison_between_two_metrics}
    \vspace{-3mm}
\end{figure}

\begin{figure}[htbp]
    \centering
    \includegraphics[width=1\linewidth]{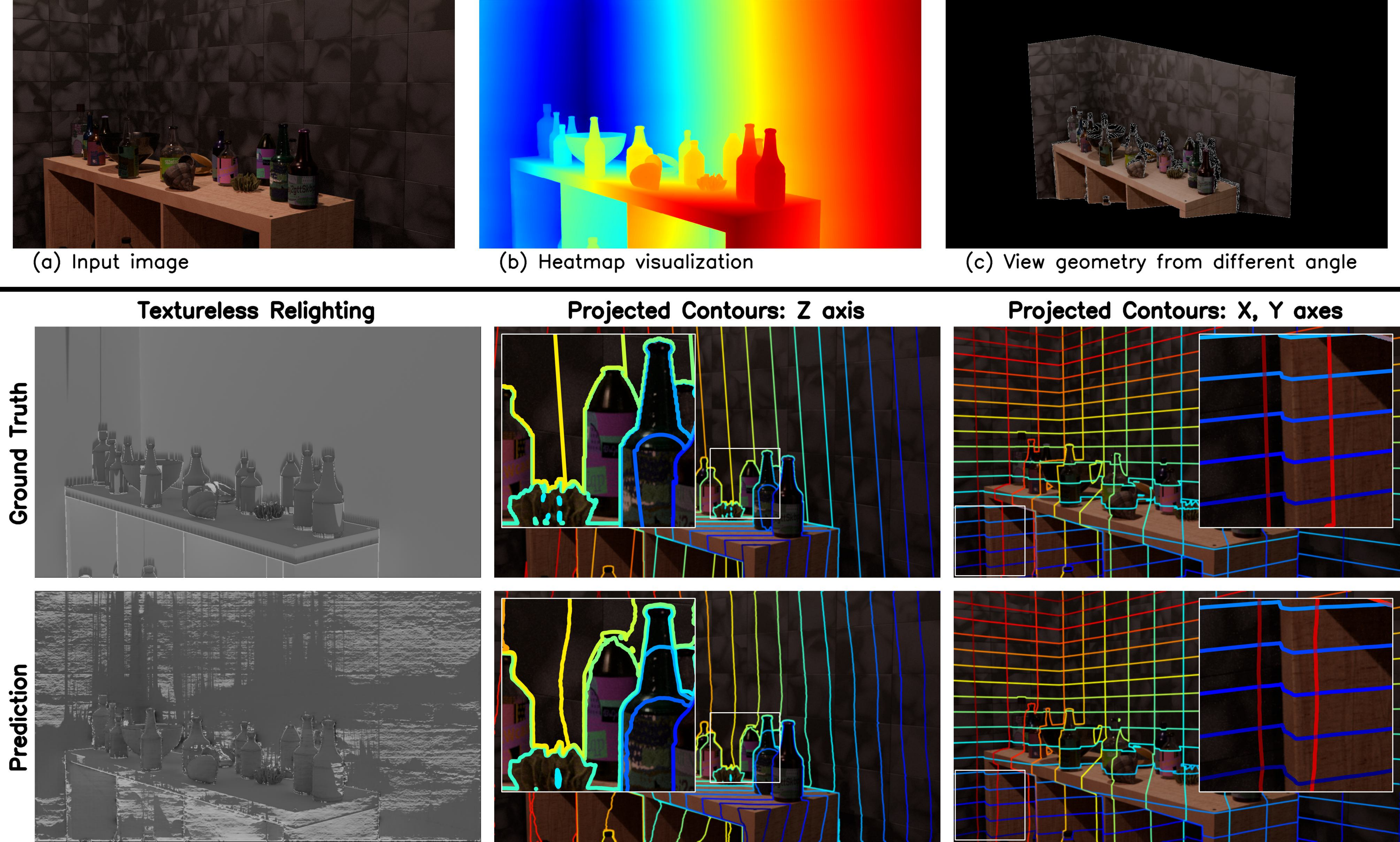}
    \caption{Top: input image and visualization of a depth using existing tools. Defects like a wavy wall are not apparent due to interference of texture. Bottom: Textureless Relighting and Projected Contours make geometric defects more apparent. Bumps on the wall are obvious.}
    \label{fig:visualization_tool}
    \vspace{-2mm}
\end{figure}

Fig.~\ref{fig:example_sensitivity_comparison_between_two_metrics} shows the exchange rate with respect to AbsRel-No Align. Comparing Boundary F1-No Align and AbsRel-No Align, the exchange rate under boundary perturbation is the highest. This is expected as Boundary F1 is designed to capture boundary sharpness which is not reflected in AbsRel and other metrics. Comparing AbsRel-Disparity Af and AbsRel-No Align, exchange rate under affine transform of disparity is the lowest. This is also expected as affine alignment on disparity can perfectly align depth of this perturbation to ground truth.

\section{Human Sensitivity}

\begin{figure*}[t]
    \centering
    \includegraphics[width=0.87\textwidth]{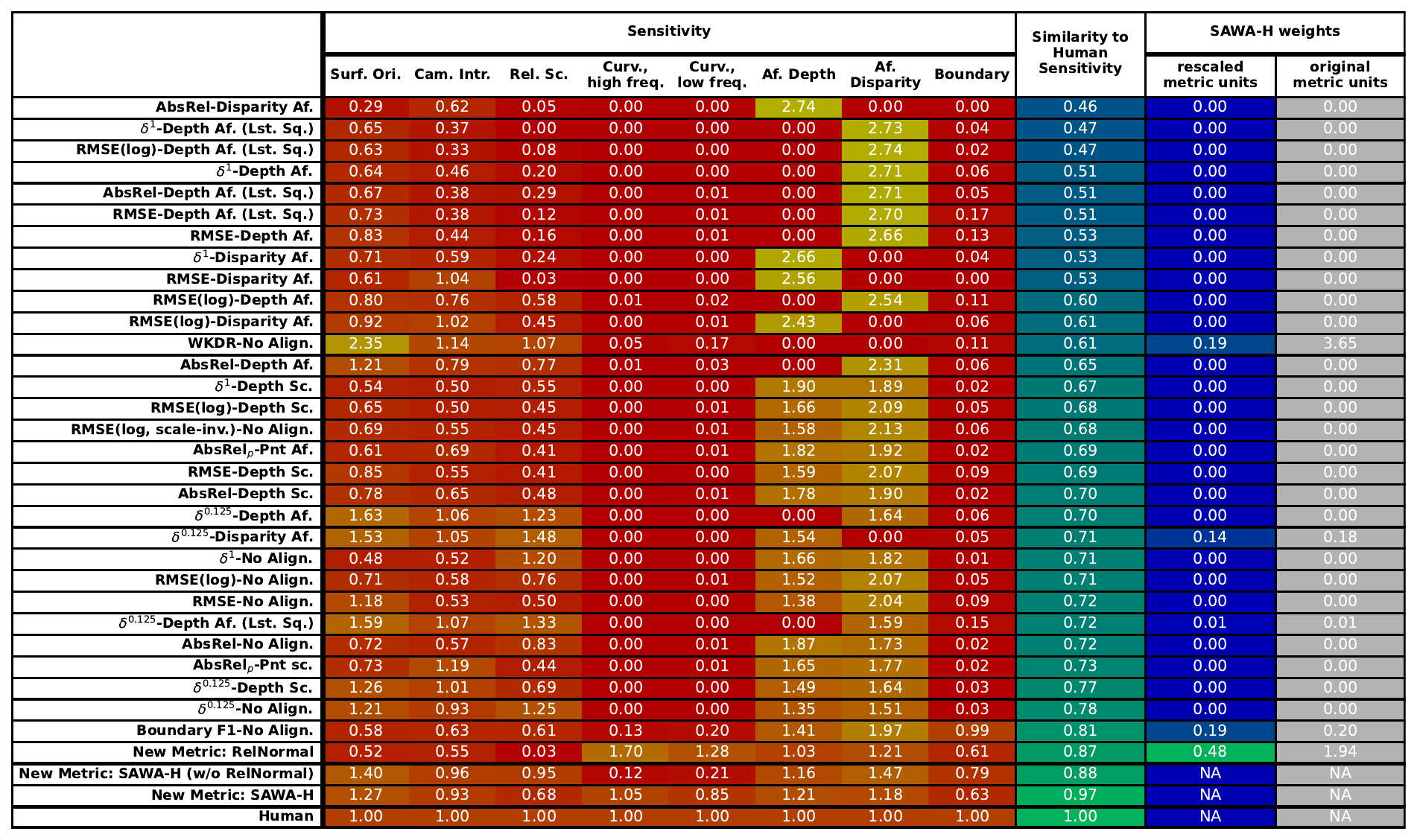}
    \caption{\textbf{Sensitivity:} Each row in red is a vector of exchange rates between a metric and human judgment, reflecting the sensitivity of a metric to different perturbations, using human judgment as a reference. Within a vector, a higher value means higher sensitivity of the metric to a perturbation, relative to humans. Human judgment (last row) against itself results in a vector of all 1s. Each vector can be arbitrarily scaled, equivalent to changing the unit of the metric; here the unit of the metric is rescaled such that the L2 norm of the vector is the same as a vector of all 1s, to facilitate comparisons across rows. All rows except the last four are existing metrics. \textbf{Similarity to human sensitivity:} cosine similarity between each sensitivity vector and the human sensitivity vector (all 1s). \textbf{SAWA-H Weights:} SAWA-H is a new, composite metric that is a weighted average of base metrics, with weights optimized to align with human sensitivity. RelNormal is a new base metric we introduce to allow better human alignment. The first column lists weights for averaging metrics under rescaled units, reflecting relative contributions invariant to the original choices of metric units; the second column lists equivalent weights for averaging metrics under their original units. } 
    \vspace{-2mm}
    \label{fig:similarity-quad}
\end{figure*}

We are particularly interested in how the sensitivity of existing metrics compares to human judgment. This is a useful question because many generative AI applications generate content for human consumption, and for such applications human judgment is highly relevant. In addition, findings of human sensitivity can reveal what is and is not important for achieving human-level visual capabilities, which might be sufficient for robotic applications. 

To measure human sensitivity,  we seek to model human judgment as a smooth non-negative function $H(x)$ of the intensity $x$ of a perturbation. This is tricky because it can be difficult for humans to produce consistent and well-calibrated numerical values, e.g., rate the depth map from 1 to 5.

Fortunately, we only care about the behavior near zero, which admits a simple solution. We show an RGB image and a depth map to a human annotator for a binary response: whether the depth map is the ground truth. The depth map is perturbed with varying intensities---we probe when a perturbation starts to become noticeable, which corresponds to a response of 1. We collect data points from multiple human annotators and scenes and fit a quadratic curve to obtain the derivative at zero. \looseness=-1

Note that unlike computer metrics, which relies on privileged access to ground truth answers, humans are not given two depth maps to compare, because two depth maps can be obviously different but indistinguishable in the terms of reconstructing the input image.  Consistent with our definition of an evaluation metric, humans are shown an RGB image and a (possibly) perturbed depth map and asked to judge whether the depth map is the correct reconstruction. This reveals human sensitivity to perturbations in the specific context of monocular depth estimation. 

\subsection{Visualization Tools}\label{sec:visualization_tool}

When measuring human sensitivity, one issue we encounter is the lack of effective depth visualization tools. Widely used visualizations include (1) a heatmap and (2) textured 3D point cloud viewed in a few new angles. Though these visualizations can illustrate coarse shape and severe deformations, they can mask many other defects. While heatmaps are widely considered insufficient, the limitation of textured point clouds is perhaps less well understood. The main issue with textured point clouds is that to the human eye, textures can create illusions of geometry that differs from the actual one, in the same way videos games use textures maps to fake geometrical details on flat surfaces. %
Consider the depth map visualized with these standard methods in Fig. \ref{fig:visualization_tool}. Is the wall flat? From only these two visualizations, it appears flat but is in fact wavy. 
To help human annotators more efficiently and effectively inspect the reconstructed 3D shape, we introduce two new visualization tools:

\noindent
 \textbf{Textureless Relighting.} When casting light to a geometry, different shape results in different patterns of shadow. Motivated by this, we develop Textureless Relighting by casting directional light to depth induced mesh. To prevent texture misleading users, the mesh is made textureless. Moreover, relighting under one direction of light is not enough, as it is hard to tell the shape of pixels in shadow. So the mesh is rendered under various lighting directions. Fig. \ref{fig:visualization_tool} Bottom Col. 1 shows an example, where bumps are easily noticeable under textureless relighting. %

\noindent
 \textbf{Projected Contours.} This visualization is made by projecting contour lines to the 3D geometry un-projected from depth. Contour lines are projected  along the direction of X,Y,Z axes. The shape of the contour lines can reveal the curvature of surfaces. For example, in Fig. \ref{fig:visualization_tool} Bottom Col. 2,3, contour lines on flat planes should be straight (first row) and wavy on bumpy surface (second row).

\vspace{-1mm}
\subsection{Sensitivity of Existing Metrics Relative to Human Judgment}
\vspace{-1mm}

\textbf{Metrics:} Using the dataset described in Sec.~\ref{sec:sensitivity_analysis}, we compare the sensitivity of 9 widely used metrics against human judgment: AbsRel, AbsRel$_p$, $\delta^{0.125}$, $\delta^1$, RMSE, RMSE (log), RMSE (log, scale invariant), WKDR, and boundary F1. Because it is common to align prediction with ground truth before computing metrics, we compute these metrics with no alignment, or with one of the following alignments if applicable: scale alignment of depth, affine alignment of depth (L1 or least square), 
affine alignment of disparity, and scale or affine alignment of point map \citep{ke2024marigold,wang2025moge}.  More details are in Appendix \ref{sec:metric_definitions}. 

\noindent
\textbf{Collecting Data of Human Judgment:} We collect human judgment on a subset of the dataset. For each nonzero perturbation intensity of each perturbation type, we randomly select 5 or more scenes and collect binary scores on the perturbed depth of those scenes. We then average the scores for each perturbation intensity, fit a quadratic curve, and compute the derivative at zero for every perturbation type. 

To detect low quality annotators who rate everything as ``not ground truth'', we randomly select 10 scenes and set the perturbation intensity to zero. This results in annotations on a total of 315 depth maps being collected. More details are in Appendix \ref{sec:appx_collect_human_judgment}.

\noindent
\textbf{Results:} Fig.~\ref{fig:similarity-quad} shows the exchange rates between every metric and human judgment under various perturbations, revealing two main findings: (1) relative to humans, existing metrics (all rows except the last four) are severely under-sensitive to the curvature perturbation; (2) humans are sensitive to affine transforms of depth or disparity, which is ignored by metrics that perform the respective affine alignment.
We use simple shapes to illustrate the theoretical causes of these observations in Appendix \ref{sec:sawa_h_diff_illustration}.

\vspace{-2mm}
\section{Sensitivity Aligned Composition}
\vspace{-2mm}

\begin{figure*}[t]
    \centering
    \includegraphics[width=0.97\linewidth]{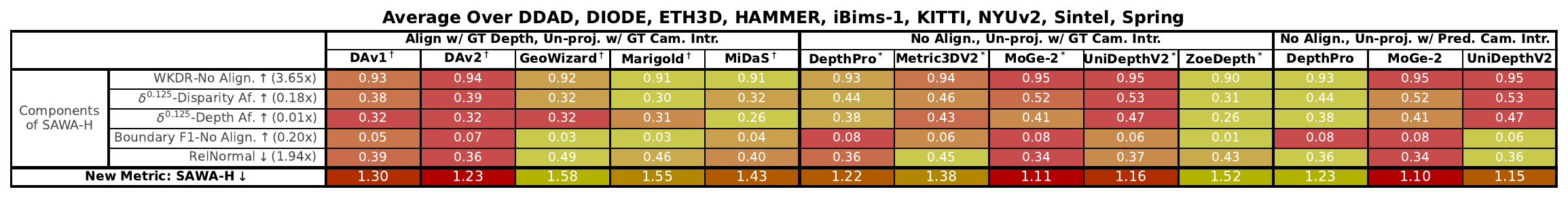}
    \vspace{-3mm}
    \caption{Performance of state-of-the-art monocular depth estimation methods under SAWA-H, averaged over 9 datasets. For component metrics of SAWA-H, the weight used to combine them to obtain SAWA-H under original metric units are listed next to metric names. \emph{Align w/ GT Depth, Un-proj. w/ GT Cam. Intr.} refers to methods whose output is first aligned with ground truth depth then un-projected using ground truth camera intrinsics; \emph{No Align., Un-proj. w/ GT Cam. Intr.} refers to methods whose output is directly un-projected using ground truth camera intrinsics; \emph{No Align., Un-proj. w/ Pred. Cam. Intr.} refers to methods whose output is directly un-projected using predicted camera intrinsics. Note that results are not directly comparable if one method is aligned with ground truth or un-projected using ground truth camera intrinsics while the other is not.} 
    \label{fig:eval_mde_methods_using_sawah}
    \vspace{-5mm}
\end{figure*}

Our sensitivity analysis of existing metrics reveals that no existing metrics align well with human judgment. How do we create a metric that does align well, if one is desired for some reason? %

We introduce Sensitivity Aligned Composition (SAC), a method to compose a set of existing metrics such that the new metric achieves a desired sensitivity profile, such as that of humans or a specific application. The idea is to combine existing metrics to form a new metric, with the combination parameters optimized such that new metric is maximally similar to a target metric (which can be a black-box oracle like human judgment) in terms of sensitivity. 

Given perturbations $P_1, \ldots, P_M$, we define the sensitivity vector of a metric A relative to a reference metric Z as a vector of exchange rates under the perturbations $\mathbf{R}(A;Z|\mathbf{P})=(R(A;Z|P_1), \ldots, R(A;Z | P_M)) \in \mathbb{R}^M$. We then define the similarity of two sensitivity vectors using cosine similarity. This is because the magnitude of the sensitivity vector does not matter and is due to the choice of the unit of the metric. 
Let $\mathbf{T}\in \mathbb{R}^M$ be a target sensitivity vector, we can optimize the parameters $w$ of a composite metric $C(w)$ to maximize the cosine similarity between $\mathbf{R}(C(w);Z|P)$ and $\mathbf{T}$.     

\noindent \textbf{Sensitivity Aligned Weighted Average (SAWA):} The simplest way to compose metrics is a non-negatively weighted average $C(w)\triangleq \sum_i w_i A_i$ of metrics $A_i$. Due to the linearity of derivative, it is easy to verify that the sensitivity vector of metric $C(w)$ relative to a reference metric $Z$ is a linear combination of the sensitivity vectors of metrics $A_i$, that is, 
    $\mathbf{R}(C(w);Z|\mathbf{P}) = \sum_i w_i \mathbf{R}(A_i;Z|\mathbf{P})$. 

Setting the reference metric to human judgment $H$ and the target sensitivity vector to $\mathbf{1}_M$ (a vector of all 1s) leads to the following optimization problem that maximizes human alignment:  
\vspace{-3mm}
\begin{equation}
\mathop{\max}\limits_{\mathbf{w}\ge 0}
\frac{\langle\sum_i w_i\mathbf{R}(A_i;H|\mathbf{P}),\mathbf{1}_M\rangle}{\|\sum_i w_i\mathbf{R}(A_i;H|\mathbf{P})\|_2\|\mathbf{1}_M\|_2}
\end{equation}
\vspace{-3mm}

\noindent
This is equivalent to a convex problem which can be solved optimally and efficiently. We refer to the resulting composite metric as SAWA-H (Sensitivity Aligned Weighted Average aligned to Human judgment). Intuitively, this optimization finds the best non-negative weights to linearly combine the red rows in Fig.~\ref{fig:similarity-quad} to maximize cosine similarity with a target vector of all 1s.  
\vspace{-5mm}
\paragraph{New metric: RelNormal}
Because all existing metrics are insensitive to curvature error, this limits how much we can align with human judgment. Intuitively, the rows of existing metrics in Fig.~\ref{fig:similarity-quad} cannot linearly combine to approximate well a vector of all 1s. To remedy this we introduce a new metric based on relative surface normals.

Similar to WKDR (\cite{zoran2015ordinalrelationship}) and boundary F1 score (\cite{bochkovskiydepthpro}), we consider the relation between patches (sets of pixels) within the depth map. Given patches $p$ and $q$ we calculate the normal of the surface described by each patch and the angle between these normals $\angle(\hat{n}_p, \hat{n}_q),\angle(n_p, n_q)$ for both ground truth and prediction. We take the average difference over many such patches,
\vspace{-2mm}
\begin{equation}
    \text{RelNormal} = \frac{1}{\pi}\cdot\frac{1}{|\mathcal{C}|}\cdot \sum_{p,q\in\mathcal{C}} \left| \angle(\hat{n}_p, \hat{n}_q) - \angle (n_p, n_q)\right|,
\end{equation}
\vspace{-3mm}

\noindent
where $\frac{1}{\pi}$ normalizes the difference to $[0,1]$.
If $p$ and $q$ lie on a continuous surface connected by a circular arc, this measures error in the degree of curvature. For $p$ and $q$ on separate but adjacent surfaces, this metric captures the local geometry (e.g. the angle between two walls in a room).

Our selection of patches is motivated by the following procedure: first uniformly select a pixel $I$ within the image, then select a pixel $J$ uniformly within a neighborhood of $I$, finally compute the error. To approximate this expectation over a multivariate uniform distribution, we use the first $m$ elements of the Sobol sequence. This provides a deterministic algorithm for computing the metric. In Appendix \ref{sec:rel_normal_sampling} we show that 1 million samples is sufficient when computing this metric on standard RGB-D datasets. We take the average over multiple scales by downsampling the image by a factor $k$ and computing the relative normals. The following results use a neighborhood radius of 32 and scale factors $[1,2,4,8]$.
We compute normals using the cross product of vectors from the top/bottom and left/right pixels adjacent to a central pixel. For datasets with noisy ground truth, it may be preferable to compute normals with a plane of best fit.

\noindent
\textbf{SAWA-H (with RelNormal)} With RelNormal, we are able to achieve substantially better human alignment, improving the cosine similarity to $0.97$ (with RelNormal) from $0.88$ (without RelNormal). Fig.~\ref{fig:similarity-quad} shows the SAWA-H weights and similarity after including RelNormal. 

Note that we do not claim that SAWA-H is a better metric, or a substitute for any existing metric. We only claim that SAWA-H aligns better with human judgment, which may or may not be desirable depending on the use case. Also, our alignment approach is not restricted to human preferences. One can create composite metrics to align with arbitrary preferences based on the desired sensitivity to perturbations.

\begin{figure}[t]
    \centering
    \includegraphics[width=1.0\linewidth]{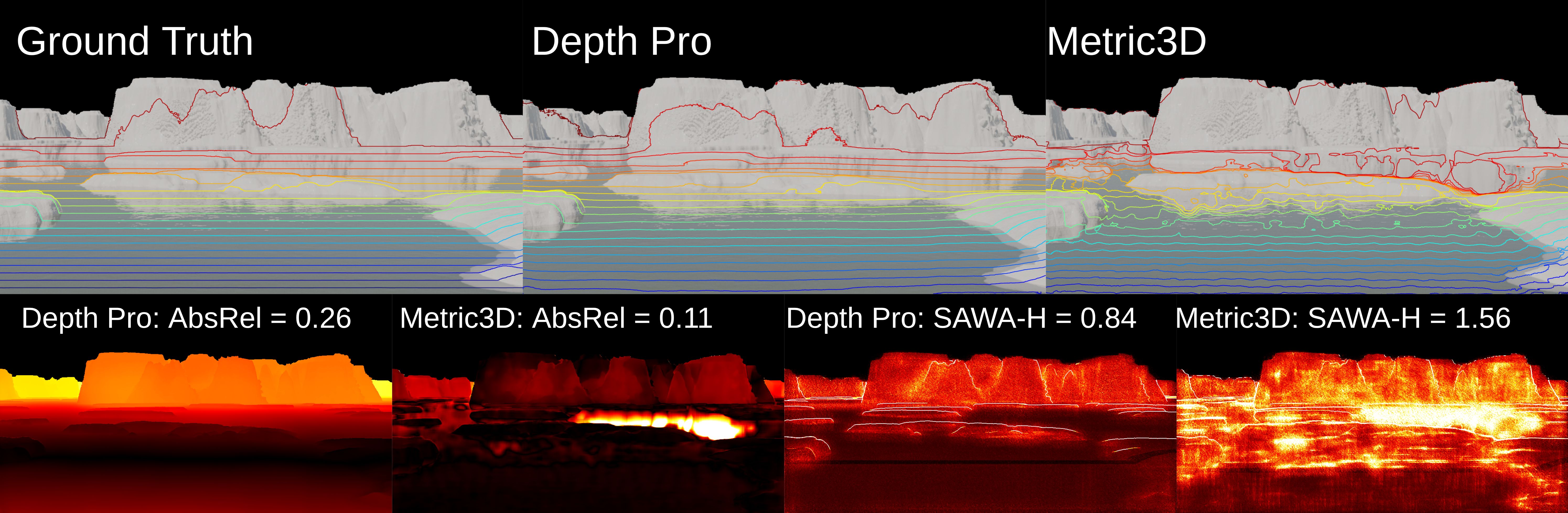}
    \vspace{-7mm}
    \caption{Disagreement between SAWA-H and AbsRel. Top row: projected contour plots of the ground truth and predicted depths. Bottom row: heatmaps for model predictions calculated with AbsRel and SAWA-H. AbsRel is computed by aligning the scale with the ground truth.} 
    \label{fig:sawa_absrel_disagreement}
    \vspace{-6mm}
\end{figure}

\noindent
\textbf{Example of SAWA-H on Real Predictions}
In some cases, the SAWA-H (with RelNormal) metric provides different rankings among depth predictions than the AbsRel metric. In Fig. \ref{fig:sawa_absrel_disagreement} we visualize one such difference for the predictions of Depth Pro and Metric3DV2 on an Infinigen scene using error maps. This is mostly straightforward, though to compute the error map for the boundary f1 score we divide the error equally among all edge pixels where the ground truth and predicted depth map disagree.
Though Metric3D achieves a smaller AbsRel error, its SAWA-H error is substantially larger. This is due to the fact that the Depth Pro model made a mistake judging the scale of the iceberg and orientation of the camera relative to ocean. Hence, the iceberg and ocean cannot both be properly aligned, so the global reconstruction is inaccurate and the AbsRel error is high. As seen in the projected contour plot, Depth Pro better approximates the local geometry and has a lower SAWA-H error score. We present more examples with predictions on Infinigen and iBims \citep{ibims} in Appendix ~\ref{sec:appx_sawah_visualization}. It is important to note that this example does \emph{not} show that SAWA-H is a better metric than AbsRel because that depends on the application; it simply illustrates that their different behaviors. 

\noindent
\textbf{Performance of MDE Methods under SAWA-H} We use SAWA-H (with RelNormal) to evaluate 10 state-of-the-art monocular depth estimation methods: DepthAnything\citep{yang2024depthanything}, DepthAnythingV2\citep{yang2024depthanythingv2}, GeoWizard\citep{fu2024geowizard}, Marigold\citep{ke2024marigold}, MiDaS\citep{birkl2023midas}, DepthPro\citep{bochkovskiydepthpro}, Metric3DV2\citep{hu2024metric3d}, MoGe-2\citep{wang2025moge2}, UniDepthV2\citep{piccinelli2025unidepthv2}, and ZoeDepth\citep{bhat2023zoedepth}. 
Evaluation is conducted on 9 commonly used benchmarks: DDAD\citep{guizilini2020DDAD}, DIODE\citep{vasiljevic2019diode}, ETH3D\citep{schops2019bad_eth3d}, HAMMER\citep{jung2023hammer}, iBims-1\citep{ibims}, KITTI\citep{uhrig2017sparsity_kitti}, NYUv2\citep{silberman2012indoor_nyuv2}, Sintel\citep{butler2012naturalistic_sintel} and Spring\citep{mehl2023spring}.

Like some existing metrics, computing SAWA-H requires not just the depth map prediction but also the camera intrinsics. We use either ground truth or predicted (if applicable) camera intrinsics. If the method does not predict metric depth (e.g. predicts affine invariant depth/disparity), we first align prediction with ground truth depth using the conventional alignment strategy (e.g. perform affine alignment of disparity for affine invariant disparity prediction) and then un-project the aligned depth. Note that evaluation results that use ground truth depth for alignment are not directly comparable to those that do not. The same holds for using ground truth camera intrinsics for back-projection. 

Fig.~\ref{fig:eval_mde_methods_using_sawah} presents metric values of each method averaged over 9 datasets. Performance of methods on each dataset are reported in Appendix~\ref{sec:appx_eval_mde_methods}. Note that Boundary F1 is one component of SAWA-H but some datasets do not provide valid depth value on pixels near boundaries, preventing SAWA-H from properly penalizing blurry boundaries on those datasets. We additionally report performance of methods averaged on datasets with accurate depth across boundaries in Appendix~\ref{sec:appx_eval_mde_methods}. 

\vspace{-3mm}
\section{Limitations and Conclusions}
\vspace{-2mm}
One limitation is that our list of perturbations, although representative, may not be exhaustive. Our dataset for sensitivity measurement may have room for improvement in diversity and coverage. These limitations may affect our numerical results such as particular sensitivity values. On the other hand, most of our contributions are on methodology and tools, and we expect them to stay useful and increasingly so, as more perturbations and better data become available.  

\section{Acknowledgments}
This work was partially supported by the National Science Foundation.

{
    \small
    \bibliographystyle{ieeenat_fullname}
    \bibliography{main}
}

\clearpage
\setcounter{page}{1}
\maketitlesupplementary

\section{Perturbation Algorithm Of Controllable Intensity}
\label{sec:perturbation_algo}
Here we provide details of algorithms to generate each type of perturbation.

\textbf{Surface Orientation Perturbation.} We first un-project ground truth depth to 3D, and compute the ground truth surface normal. Then we rotate every ground truth surface normal by the same amount. Perturbation intensity is defined as the magnitude of rotation (i.e. rotated by $s$ degrees). Using the rotated surface normal, we compute the log depth gradient such that un-projection of neighboring pixels in 3D following the log depth gradient is perpendicular to the rotated surface normal. Then we solve the optimization problem of minimizing L2 distance between this target log depth gradient and the log depth gradient of perturbed depth, where the perturbed depth is treated as unknown variable. The optimization problem is solved by the conjugate gradient algorithm, with ground truth depth as initialization. We ignore constraints across occlusion boundaries to avoid them dominating the optimization objective, which leads to poor shape of perturbed depth.

\textbf{Camera Intrinsics Perturbation.} In this perturbation, focal length is perturbed by $s$ times, $(s\ge 1)$, i.e. $f'=sf^*$, where $f'$ and $f^*$ are predicted and ground truth focal length. $s$ controls perturbation intensity. Similar to surface orientation perturbation, we compute the ground truth surface normal. From the ground truth surface normals, we again compute the log depth gradient, but under the {\it perturbed} focal length. Then, we solve the optimization of minimizing L2 distance between this target log depth gradient computed from ground truth surface normal and the log depth gradient of perturbed depth similarly. 

\textbf{Relative Scale Perturbation.} We partition pixels to 3 sets, $S_{\text{near}},S_{\text{between}},S_{\text{far}}$. Depth of $S_{\text{near}}$ remains unchanged. Depth of $S_{\text{far}}$ is scaled by $s(s\ge1)$, where $s$ is the parameter to control perturbation magnitude. Depth of $(i,j)\in S_{\text{between}}$ is scaled by $1+(s-1)\cdot \frac{D^*_{i,j}-d_{l}}{d_{r}-d_{l}}$, where $D^*_{i,j}$ denotes the ground truth depth, and $d_l$ and $d_r$ denotes the depth of farthest pixel in $S_{\text{near}}$ and the nearest pixel in $S_{\text{far}}$ in  ground truth. To partition pixels, if the image can be split into two regions by occlusion boundaries, we set $S_{\text{near}}$ to be the region closer to the camera, $S_{\text{far}}$ to be the region farther to the camera, and $S_{\text{between}}$ to be empty. For images that cannot be split into such two regions, we find $d_l,d_r(d_l<d_r)$ that minimizes $d_r/d_l$, such that there are 5\% pixels with depth value in between, at least 30\% pixels with depth value $<d_l$, and at least 30\% pixels with depth value $>d_r$. $S_{\text{near}},S_{\text{between}},S_{\text{far}}$ are then set to be pixels with depth value $\le d_l,\in[d_l,d_r],$ and $\ge d_r$. %

\textbf{Curvature Perturbation.} In this perturbation, we first generate a HxW noise map $K\in\mathbb{R}^{H\times W}$, where $K_{i,j}$ are independently uniformly sampled from $[1-s,1+s]$. Here $s(s\ge 0)$ controls perturbation magnitude. Then the perturbed depth, $D'$, is generated by $D'\equiv D^*\otimes\text{clip}(\text{Gaussian Smooth}(K,\sigma),\min=0.1)$, where $D^*$ is ground truth depth, $\otimes$ denotes element-wise product, and $\sigma$ controls perturbation frequency. We choose $\sigma=1$ for high frequency perturbation, and $\sigma=10$ for low frequency.

\textbf{Affine Transform Perturbation.} We study affine transform of depth and disparity. For affine transform of depth, perturbed depth $D'$ follows $D'\equiv \frac{1}{s}D^*+\text{median}(D^*)-\text{median}(\frac{1}{s}D^*)$, where $s\ge 1$ controls perturbation magnitude. And for affine transform of disparity, perturbed depth $D'$ follows $\frac{1}{D'}\equiv \frac{1}{s}\frac{1}{D^*}+\text{median}(\frac{1}{D^*})-\text{median}(\frac{1}{s}\frac{1}{D^*})$, where $s\ge 1$ controls perturbation magnitude.

\textbf{Boundary Perturbation.} We apply mean filter of patch size $2s+1$, where $s\ge 0$ controls perturbation magnitude. For each pixel, the perturbed depth is further clipped by 0.7x and 1.3x of ground truth depth.

\section{More on Collecting Data of Human Judgment}
\label{sec:appx_collect_human_judgment}

It may be difficult to notice some artifacts of depth map from a single type of visualization, e.g. Fig.~\ref{fig:visualization_tool} Top, so we present annotators with multiple visualizations of a depth map. In particular, we provide visualizations of viewing geometry generated from different angles (Fig.~\ref{fig:visualization_tool}(c)), textureless relighting (Sec.~\ref{sec:visualization_tool}), and projected contours (Sec.~\ref{sec:visualization_tool}).

We adopt the following strategies to control annotation quality:
\begin{itemize}[leftmargin=1em, itemsep=1pt]
\item  To ensure that annotators examine all types of visualization, each time, we only show one image and one type of visualization of depth. A depth map is annotated as the ground truth depth if it is annotated as the ground truth depth under every type of visualization. Each visualization is examined independently and we do not reveal whether two visualizations are from the same depth map. 
\item Among the depth maps, 20 of them are gold-standard depth maps for quality control. An annotator's annotations are rejected if the accuracy on the gold-standard depth maps is below 70\%.
\end{itemize}

\section{Metric Definitions}
\label{sec:metric_definitions}

Here we provide definitions for the metrics used in Fig.~\ref{fig:similarity-quad}. These are all well-known or variants of well-known metrics.

For a ground truth depth map $z$ and a predicted $\hat{z}$, AbsRel is defined as the average of $|z_i - \hat{z_i}| / z_i$. Following \cite{wang2025moge}, we define AbsRel$_p$ on point maps $p \in \mathbb{R}^{H \times W \times 3}$ by the average of $\left\| p_i - \hat{p}_i\right\| / \|p_i\|$.

$\delta^1$ is defined as the fraction of pixels such that $\max\left\{d_i / \hat{d}_i, \hat{d}_i / d_i\right\} < 1.25$. We define a stricter metric $\delta^{0.125}$ by the fraction of pixels such that $\max\left\{d_i/\hat{d}_i, \hat{d}_i/d_i\right\} < 1.25^{0.125}$.

RMSE and RMSE (log) follow standard definitions. RMSE (log, scale invariant) is defined as 
\[
\text{Log RMSE SI} = \sqrt{\frac1n\sum_{i=1}^n(\log z_i - \log \hat{z_i} + \alpha)^2} %
\]
where $\alpha = \frac1n\sum_i (\log \hat{z}_i - \log z_i)$. 

This follows the definitions in \cite{ke2024marigold,eigen2014depth}.

Finally, WKDR follows the definition given in \cite{yin2021learning} based on relative depth relationships between pairs of pixels. Boundary F1 score follows the definition in \cite{bochkovskiydepthpro} to compute an F1 score over the edges in the predicted and ground truth depth maps.

The affine and scale alignment procedure used throughout is identical to MoGe (\cite{wang2025moge}). Notably, affine alignment for the point cloud metric AbsRel$_p$ is not the same as affine alignment of the depth. ``Lst. Sq.'' denotes least squares alignment on depth, as performed by Marigold (\cite{ke2024marigold}). 

\section{Causes of Different Behaviors Between Human Judgment and Existing Metrics}
\label{sec:sawa_h_diff_illustration}
\begin{figure*}[t]
    \centering
    \hspace{0.03\textwidth}
    \begin{subfigure}[t]{0.28\textwidth}
        \centering
        \includegraphics[width=\textwidth]{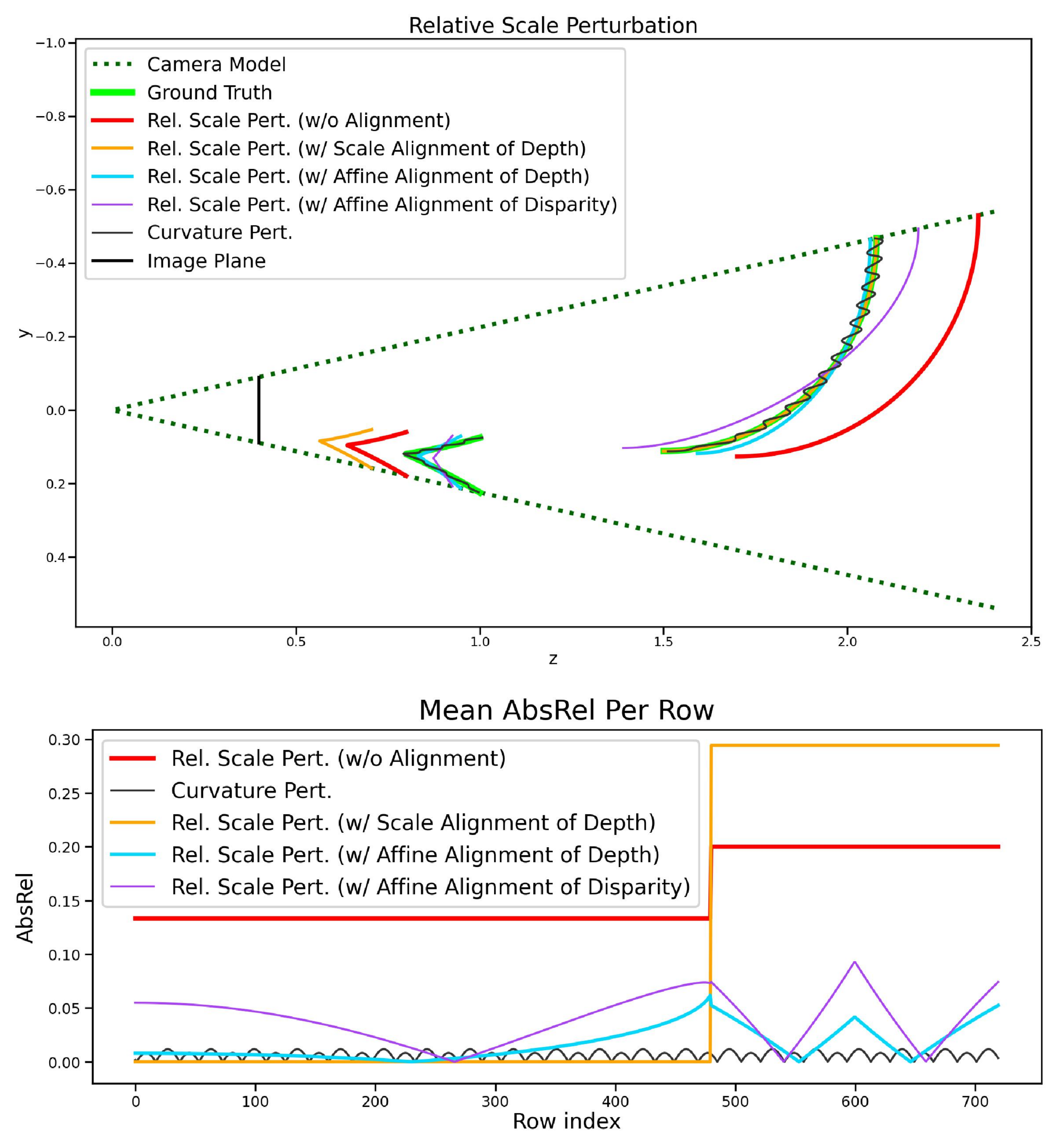}
    \end{subfigure}
    \hfill
    \begin{subfigure}[t]{0.28\textwidth}
        \centering
        \includegraphics[width=\textwidth]{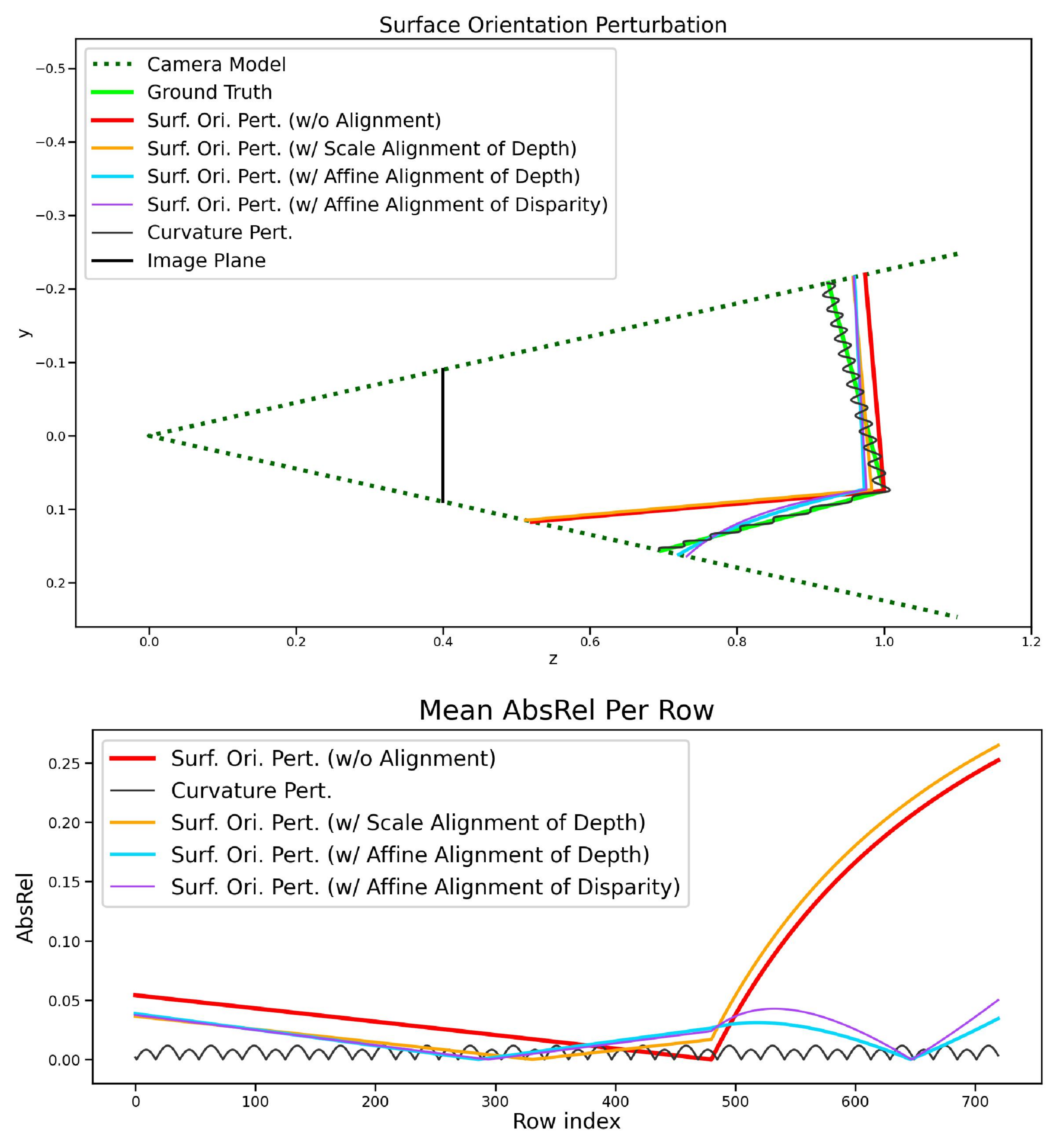}
    \end{subfigure}
    \hfill
    \begin{subfigure}[t]{0.27\textwidth}
        \centering
        \includegraphics[width=\textwidth]{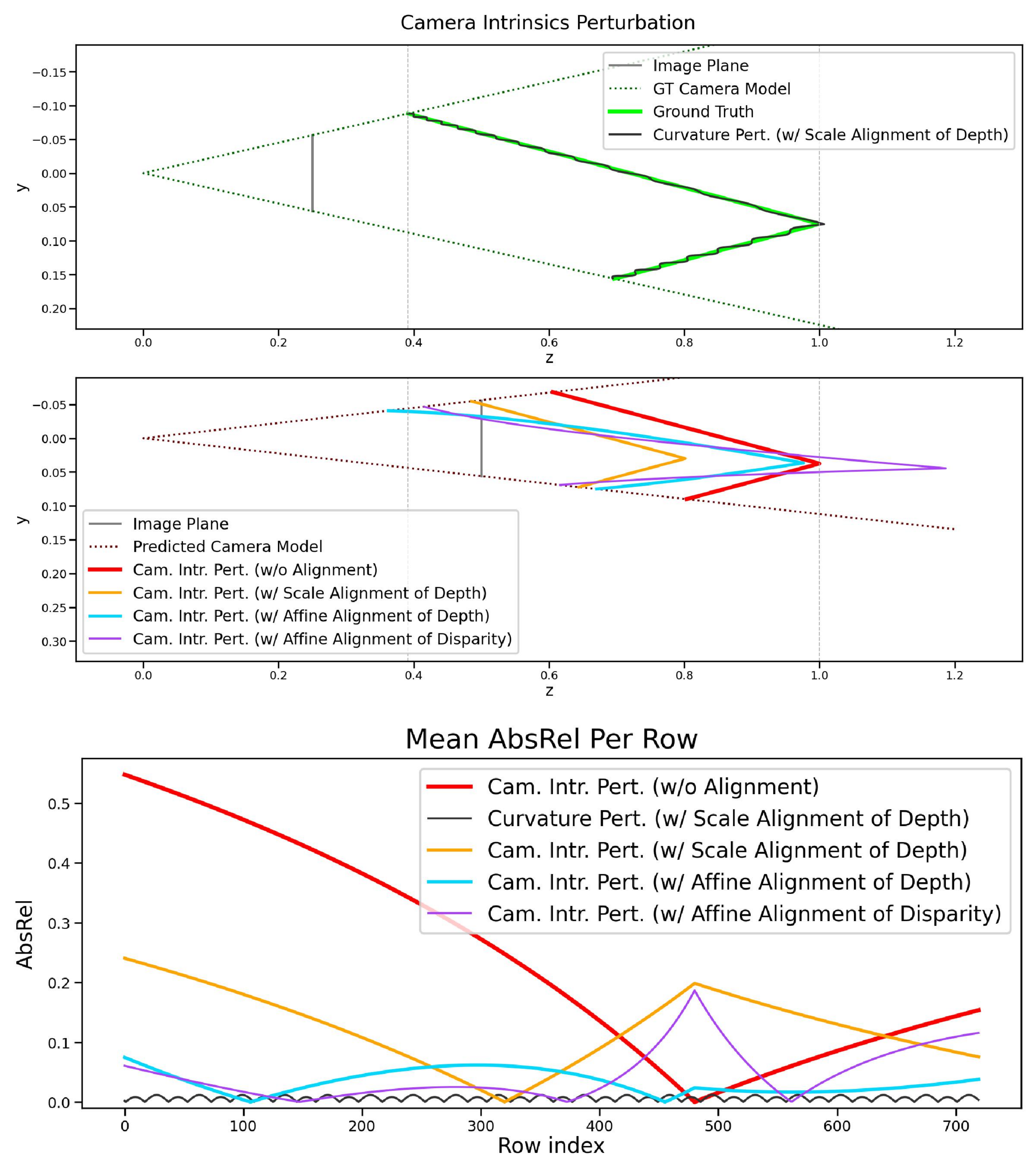}
    \end{subfigure}
    \hspace{0.03\textwidth}
    \caption{Top and Mid: side views of ground truth and different perturbations. Relative scale, surface orientation, and camera intrinsics perturbations (red) have good shape, but comparing with curvature perturbations (black), they have poor AbsRel, even under alignments (orange, blue, and purple). Bottom: mean AbsRel of each row. Curvature perturbation has much lower AbsRel.}
    \label{fig:extreme_curvature_error}
    \vspace{-2mm}
\end{figure*}

Fig.~\ref{fig:extreme_curvature_error} shows instances of relative scale, surface orientation, camera intrinsics, and curvature perturbations. On each scene, instance of curvature perturbation has much lower (better) AbsRel score, but its shape is very different from ground truth.

\begin{figure}
    \centering
    \includegraphics[width=0.8\linewidth]{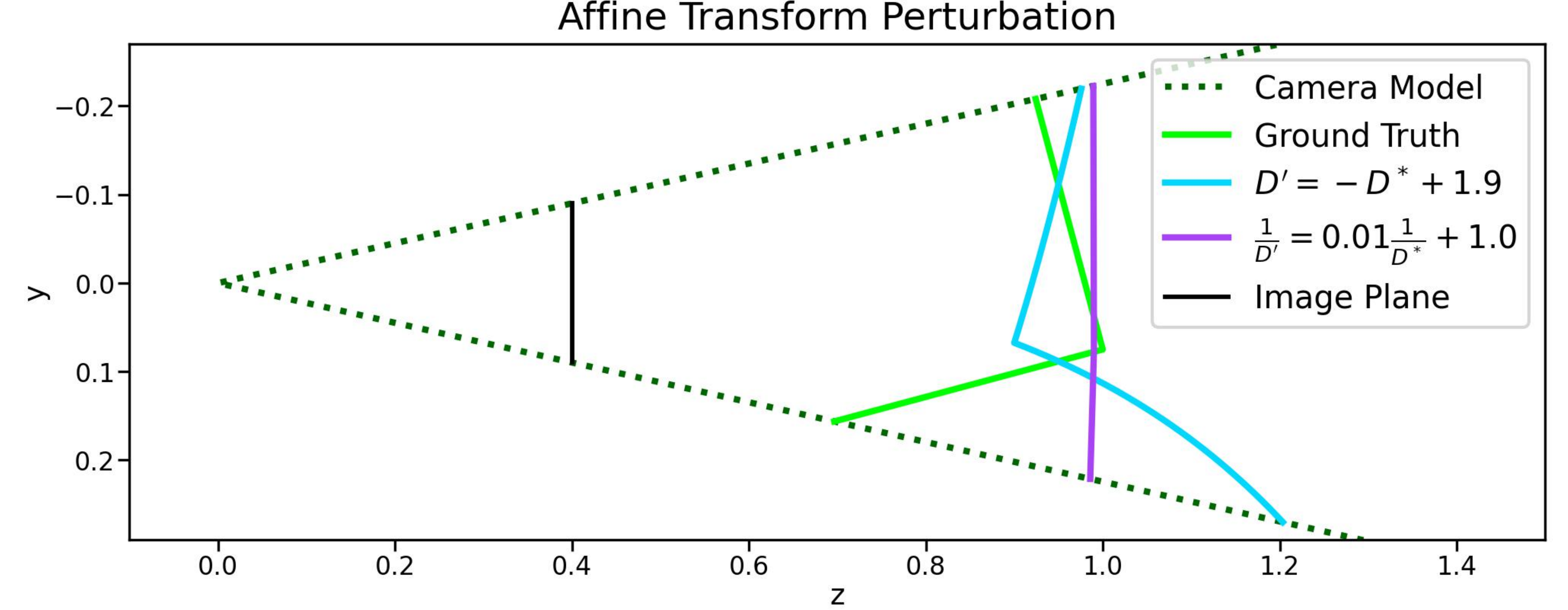}
    \caption{Side view of ground truth (green) and affine transform perturbations. Extreme affine transform parameters flatten geometry (purple) or change relative depth (blue).}
    \label{fig:affine_alignment_illustration}
    \vspace{-2mm}
\end{figure}

Fig.~\ref{fig:affine_alignment_illustration} shows instances of affine transform perturbations. The shape of these instances is distorted a lot, but they achieve a perfect score when the respective type of affine alignment is applied during evaluation.

\section{Relative Normal}
\label{sec:rel_normal_sampling}
\subsection{Sampling Procedure}
\begin{figure}[h]
    \centering
    \includegraphics[width=0.75\linewidth]{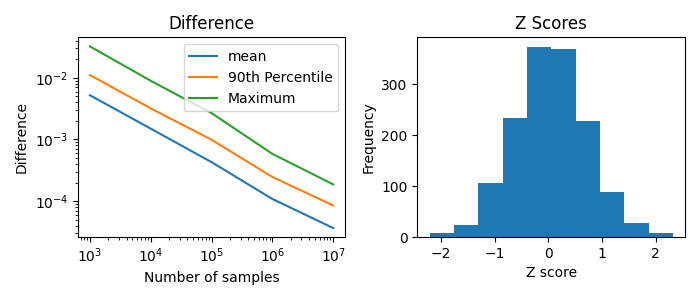}
    \caption{Comparison between deterministic and random sampling algorithm. The left plot shows the difference between the deterministic algorithm with $n$ samples and the random algorithm with $10^8$ samples. The maximum difference with 1M samples is $5.84 \times 10^{-4}$ while the mean difference is $1.08 \times 10^{-4}$. The right plot shows the z scores of the deterministic algorithm with 1M samples, compared to a distribution of the randomized algorithm with 1M samples computed with 30 different seeds.}
    \label{fig:random_sampling}
\end{figure}
When computing the relative normal metric, we wish to ensure that our deterministic sampling technique accurately approximates the true mean of the distribution. In Fig.~\ref{fig:random_sampling} we display the difference between the metric computed with the deterministic algorithm and the metric computed with $10^8$ random samples. We perform the error computations using predictions from Depth Pro (\cite{bochkovskiydepthpro}), UniDepthV2 (\cite{piccinelli2025unidepthv2universalmonocularmetric}), MoGeV1 (\cite{wang2025moge}), Metric3DV2 (\cite{hu2024metric3d}), and MoGeV2 (\cite{wang2025moge2}) on 100 images from the iBims dataset and 100 images from the Virtual KITTI dataset \cite{ibims,vkitti}. The maximum error $5.84 \times 10^{-4}$ and average error of $1.08 \times 10^{-4}$ is sufficiently small to justify using $10^6$ samples. We also analyze the $z$-score of the deterministic computation compared to random sampling. This suggests that using the Sobol sequence does not introduce unexpected irregularities. 

\section{SAWA-H Comparison Examples}
\label{sec:appx_sawah_visualization}
\begin{figure}
    \centering
    \includegraphics[width=1\linewidth]{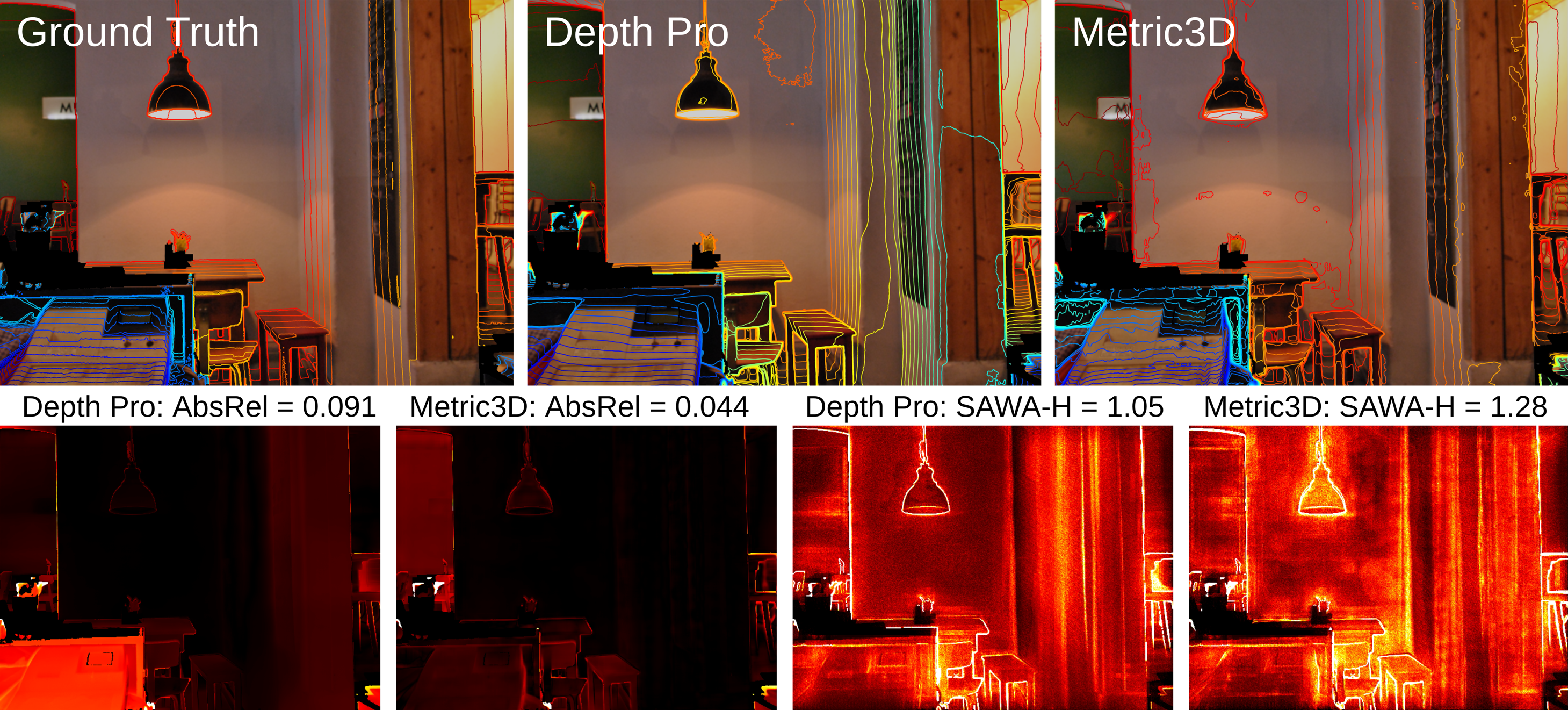}
    \includegraphics[width=1\linewidth]{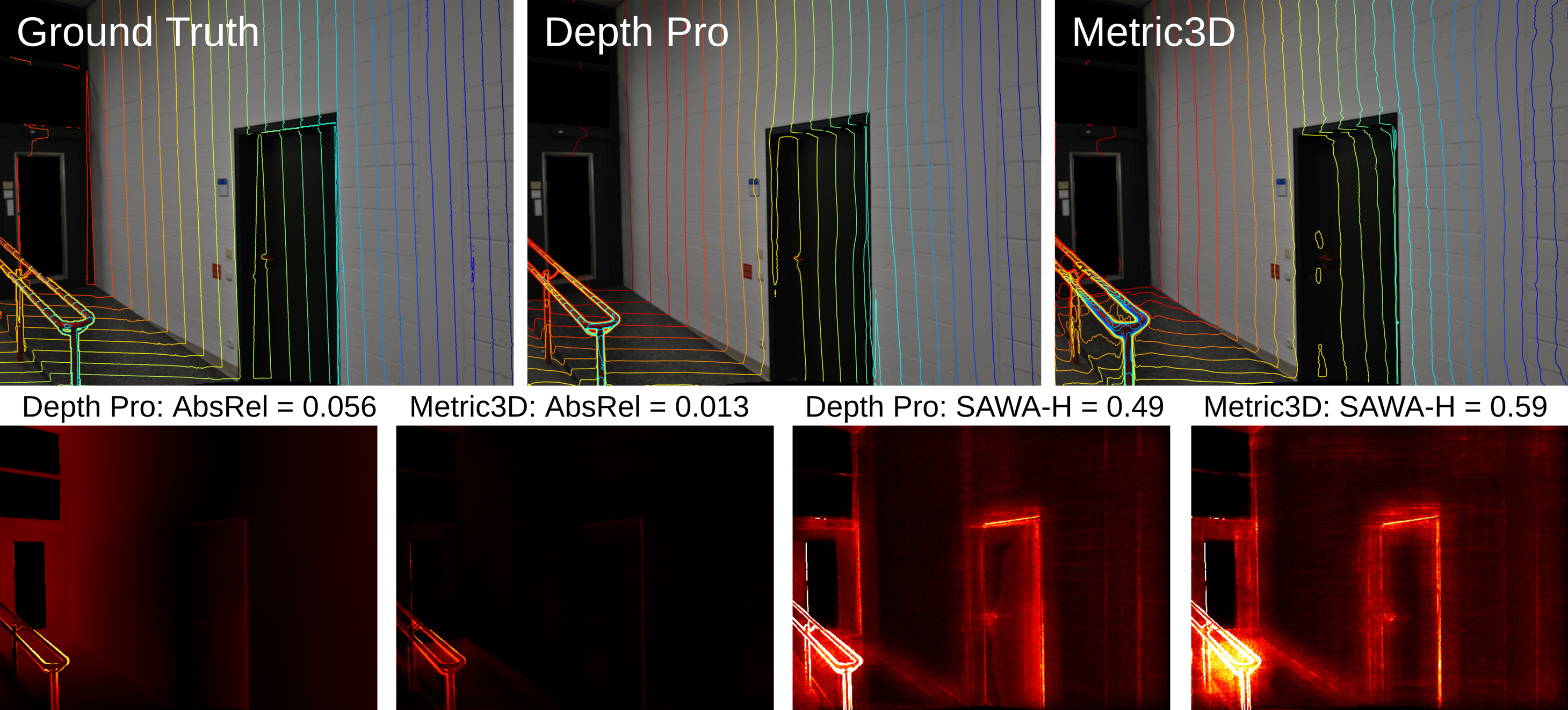}
    \includegraphics[width=1\linewidth]{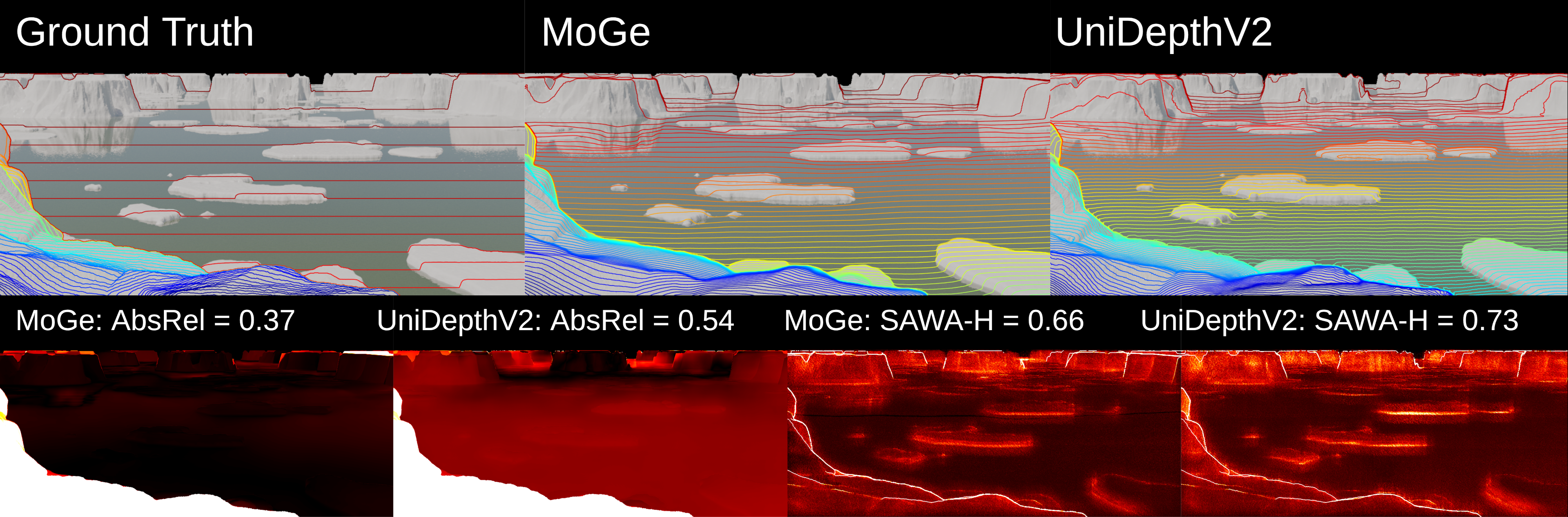}
    \caption{Additional comparisons between SAWA-H and AbsRel. The top row displays contour plots of the ground truth and predicted depths. The bottom row displays heatmaps for model predictions calculated with AbsRel and SAWA-H. AbsRel is computed by aligning the scale with the ground truth.}
    \label{fig:additional_examples}
\end{figure}

Fig.~\ref{fig:additional_examples} shows more qualitative examples of predictions evaluated by SAWA-H (with RelNormal) and AbsRel.

\section{Dataset Gallery}
\label{sec:dataset_example}
\begin{figure}[h]
    \centering
    \includegraphics[width=\linewidth]{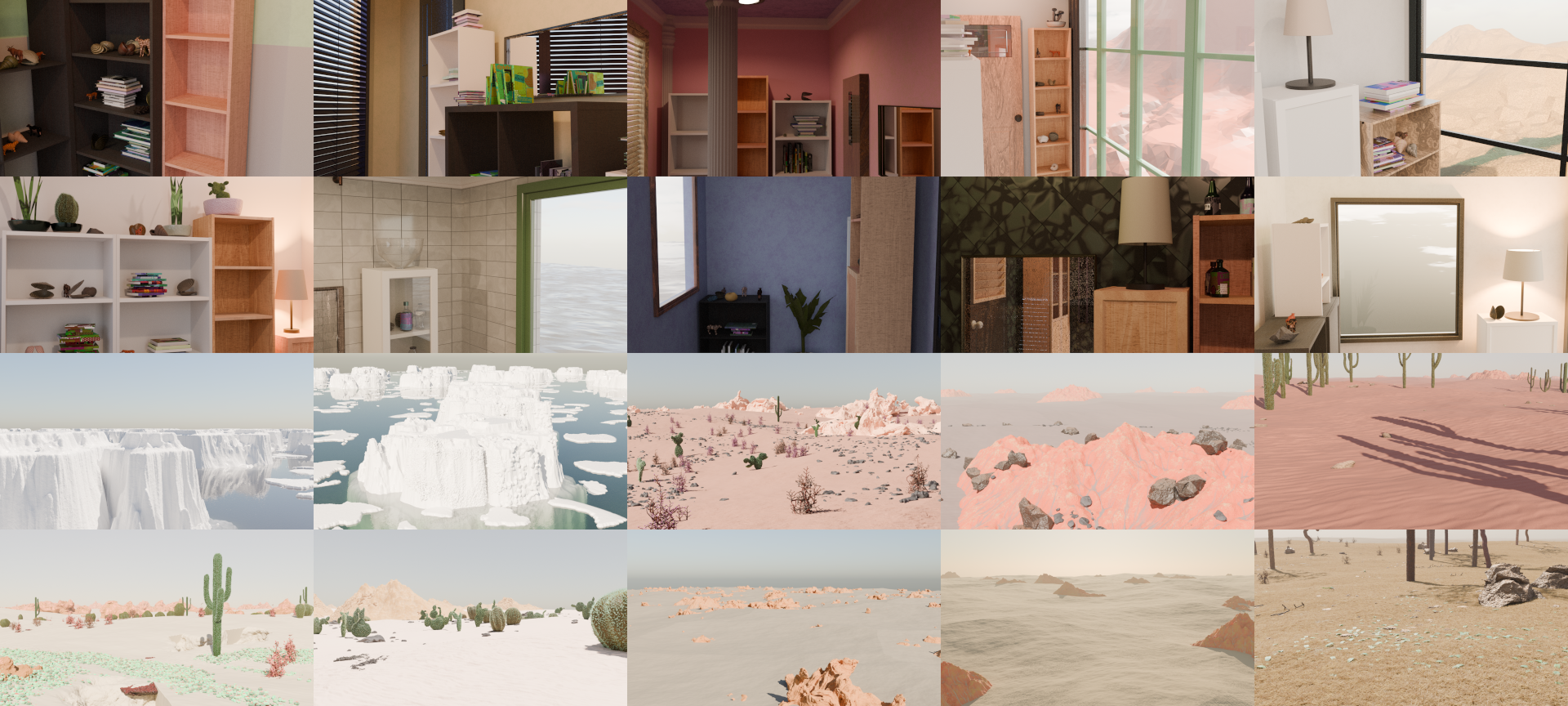}
    \caption{Sample of 20 Images from the Sensitivity Dataset}
    \label{fig:gallary_of_dataset}
\end{figure}

\begin{figure}[h]
    \centering
    \includegraphics[width=\linewidth]{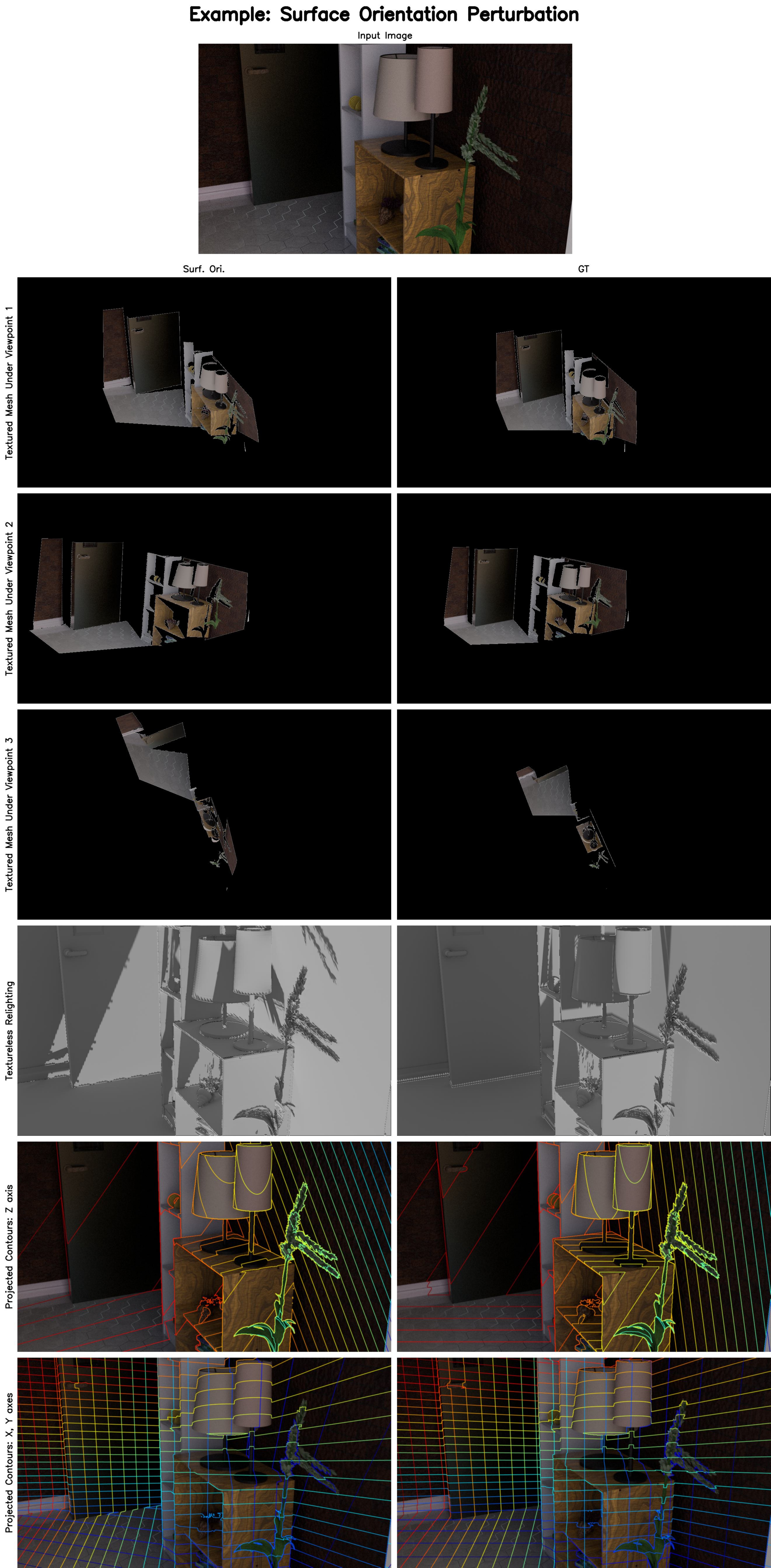}
    \caption{Example of Surface Orientation Perturbation. In visualization of textured mesh, images on the same row are captured under the same camera pose and using ground truth camera intrinsics. The geometry is perturbed by rotating $15^\circ$ along the axis of $(0.25, 0.51, -0.82)$. The orientation difference is noticeable in Projected Contours: Z axis.}
    \label{fig:real_perturbation_example_surf_ori}
\end{figure}

\begin{figure}[h]
    \centering
    \includegraphics[width=\linewidth]{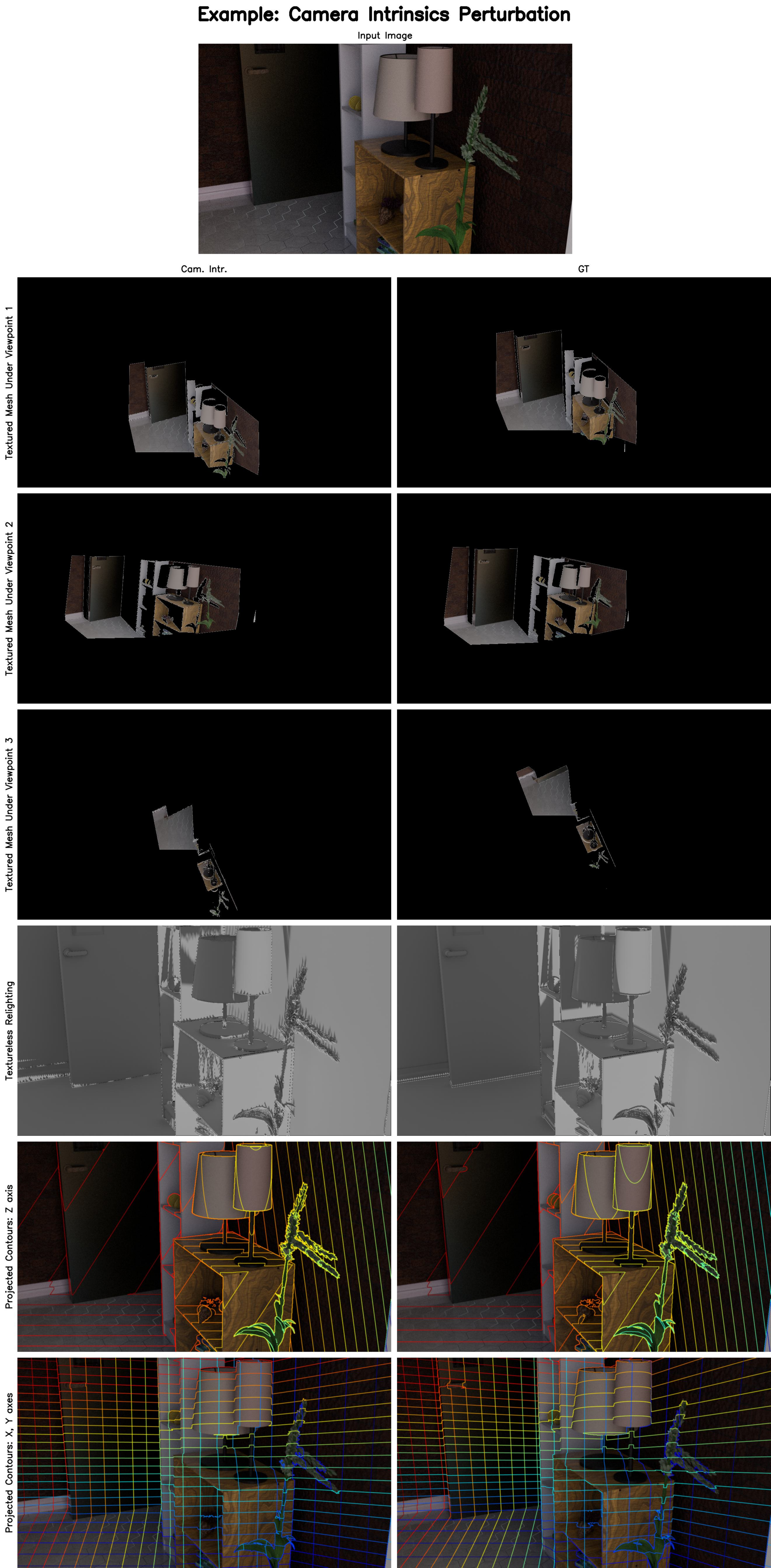}
    \caption{Example of Camera Intrinsics Perturbation. In visualization of textured mesh, images on the same row are captured under the same camera pose. Perturbed camera intrinsics are used to visualize perturbed geometry, and ground truth camera intrinsics are used for ground truth geometry. Focal length is perturbed to be 2x of ground truth focal length. To maintain similar geometry, depth range increases. This can be noticed in the textured mesh under different viewpoints.}
    \label{fig:real_perturbation_example_cam_intr}
\end{figure}

\begin{figure}[h]
    \centering
    \includegraphics[width=\linewidth]{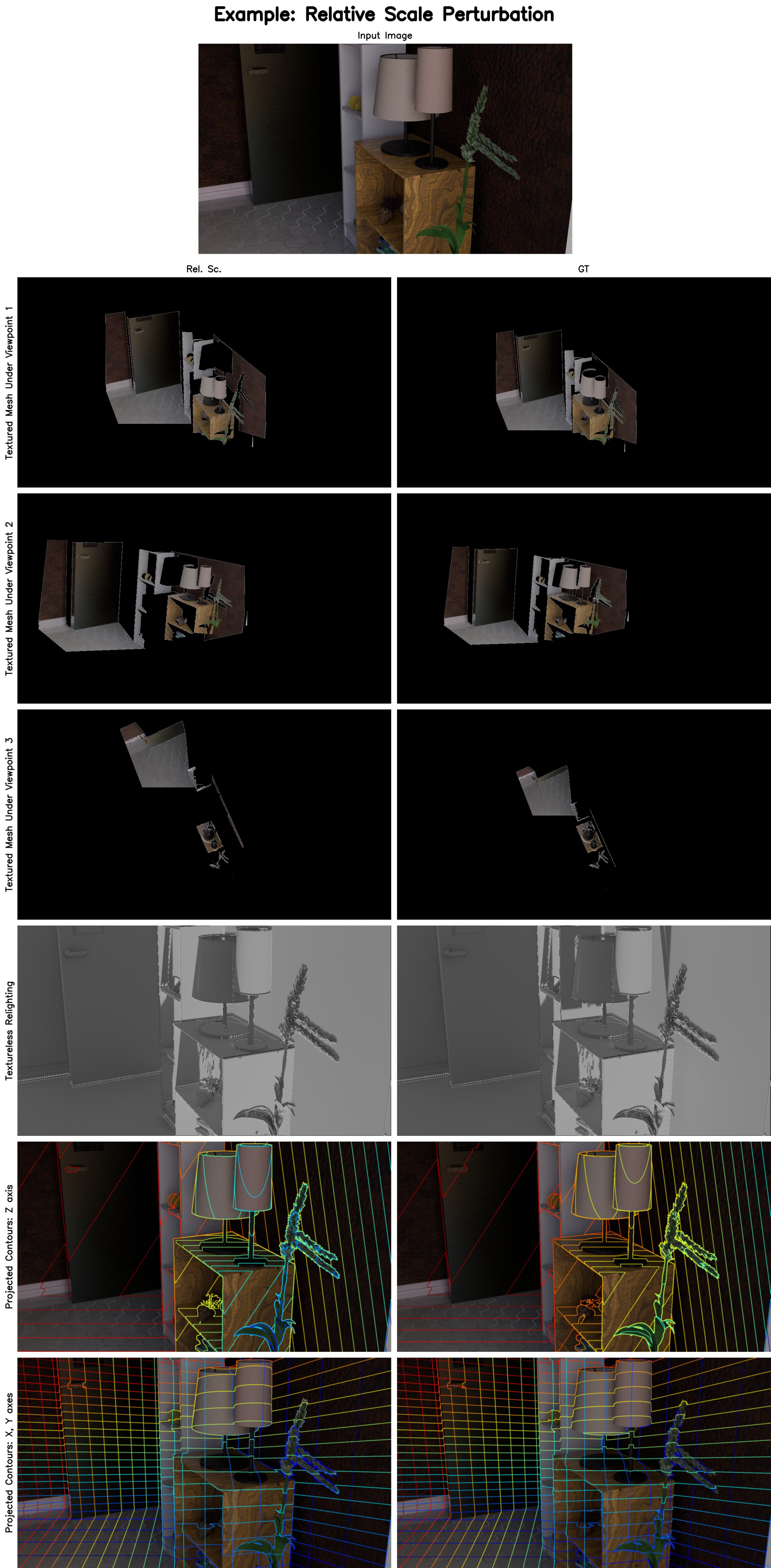}
    \caption{Example of Relative Scale Perturbation. In visualization of the textured mesh, images on the same row are captured under the same camera pose using ground truth camera intrinsics. Relative scale between the cabinet and the wall is perturbed.}
    \label{fig:real_perturbation_example_rel_sc}
\end{figure}

\begin{figure}[h]
    \centering
    \includegraphics[width=\linewidth]{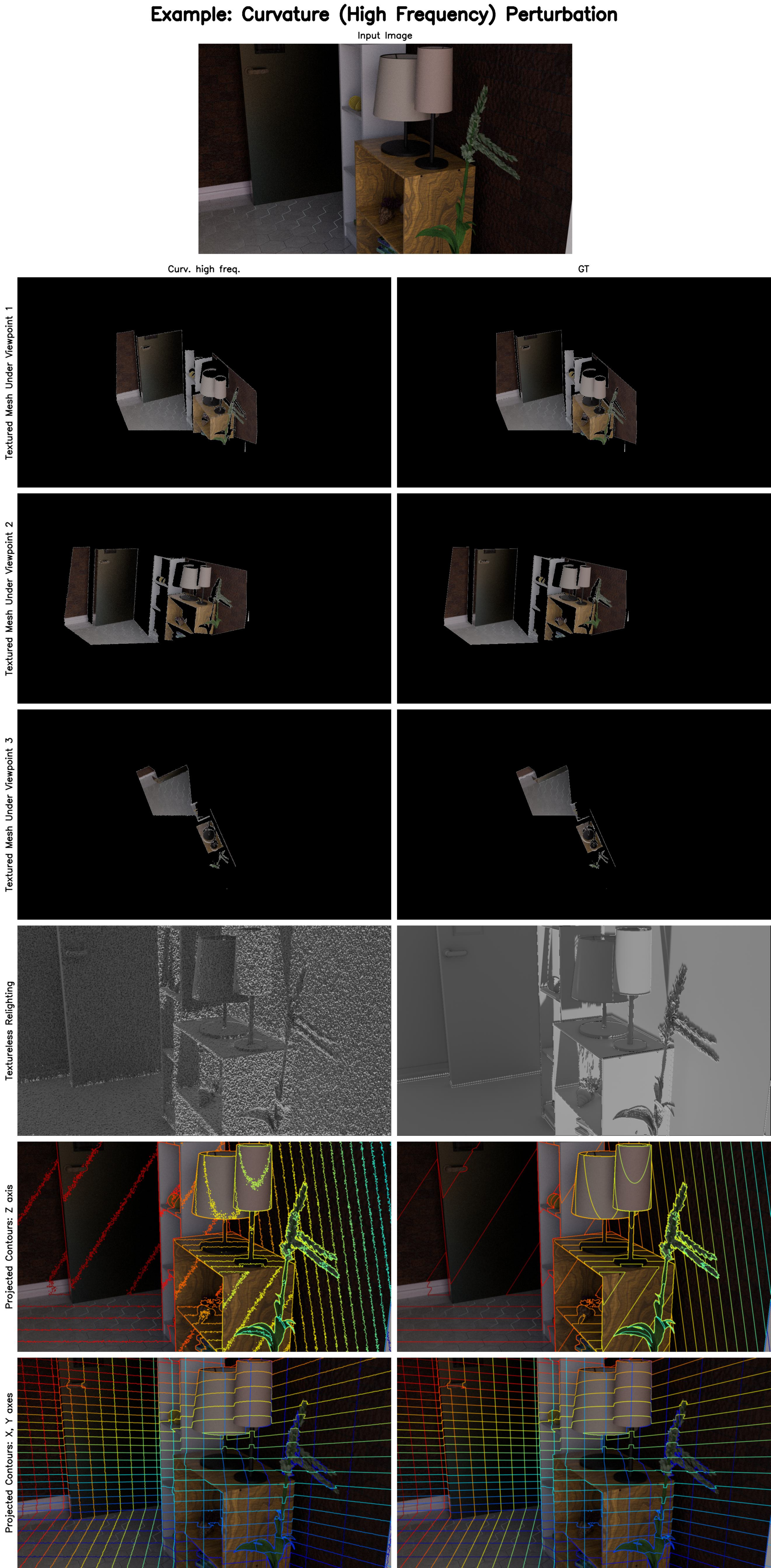}
    \caption{Example of Curvature (High Frequency) Perturbation. In visualization of the textured mesh, images on the same row are captured under the same camera pose using ground truth camera intrinsics. There are many high frequency spikes in the perturbed geometry, which can be easily observed in Textureless Relighting and Projected Contours.}
    \label{fig:real_perturbation_example_curv_high_freq}
\end{figure}

\begin{figure}[h]
    \centering
    \includegraphics[width=\linewidth]{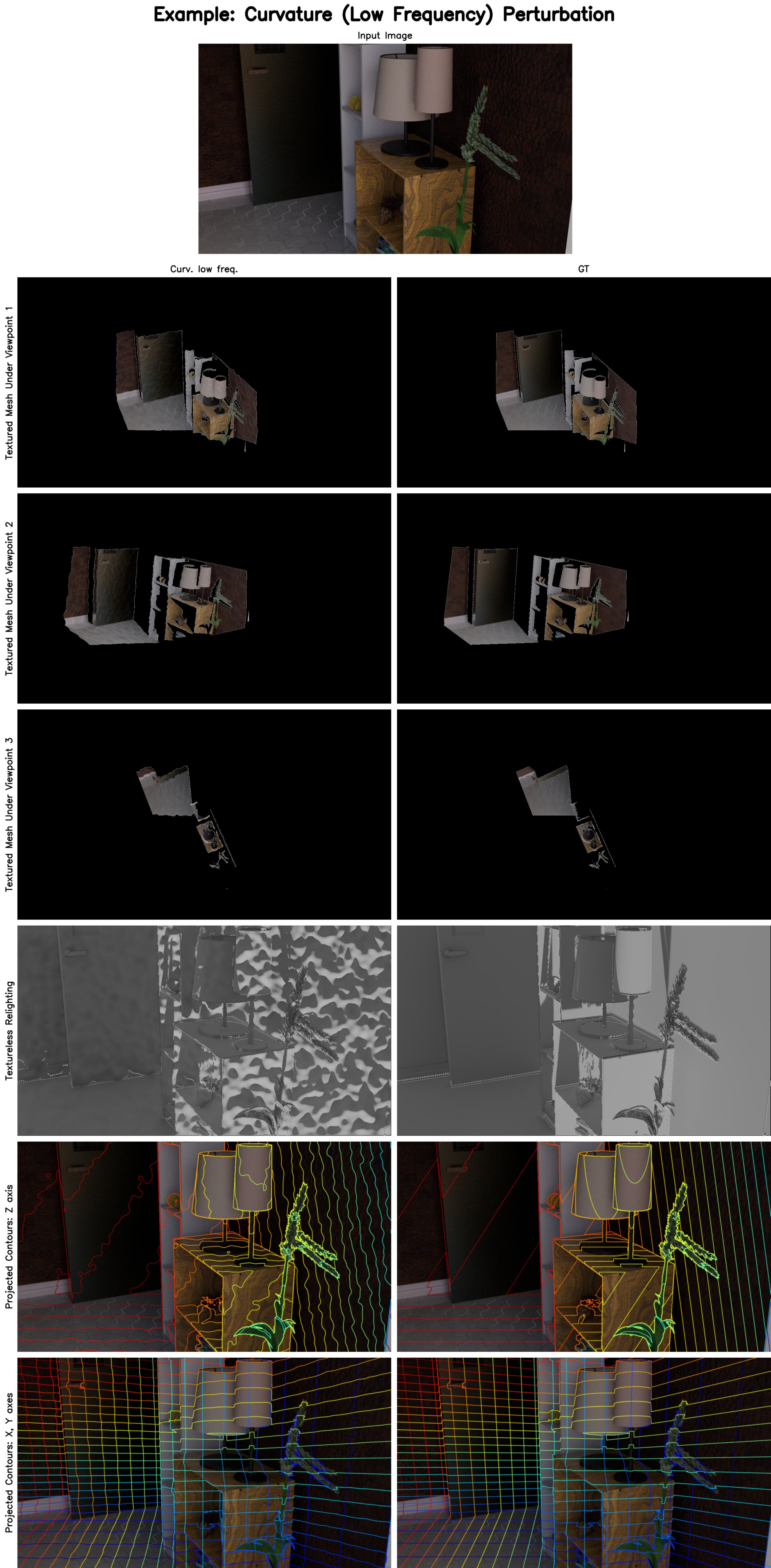}
    \caption{Example of Curvature (Low Frequency) Perturbation. In visualization of textured mesh, images on the same row are captured under the same camera pose and using ground truth camera intrinsics. There are many bumps in the perturbed geometry, which can be easily observed in Textureless Relighting and Projected Contours.}
    \label{fig:real_perturbation_example_curv_low_freq}
\end{figure}

\begin{figure}[h]
    \centering
    \includegraphics[width=\linewidth]{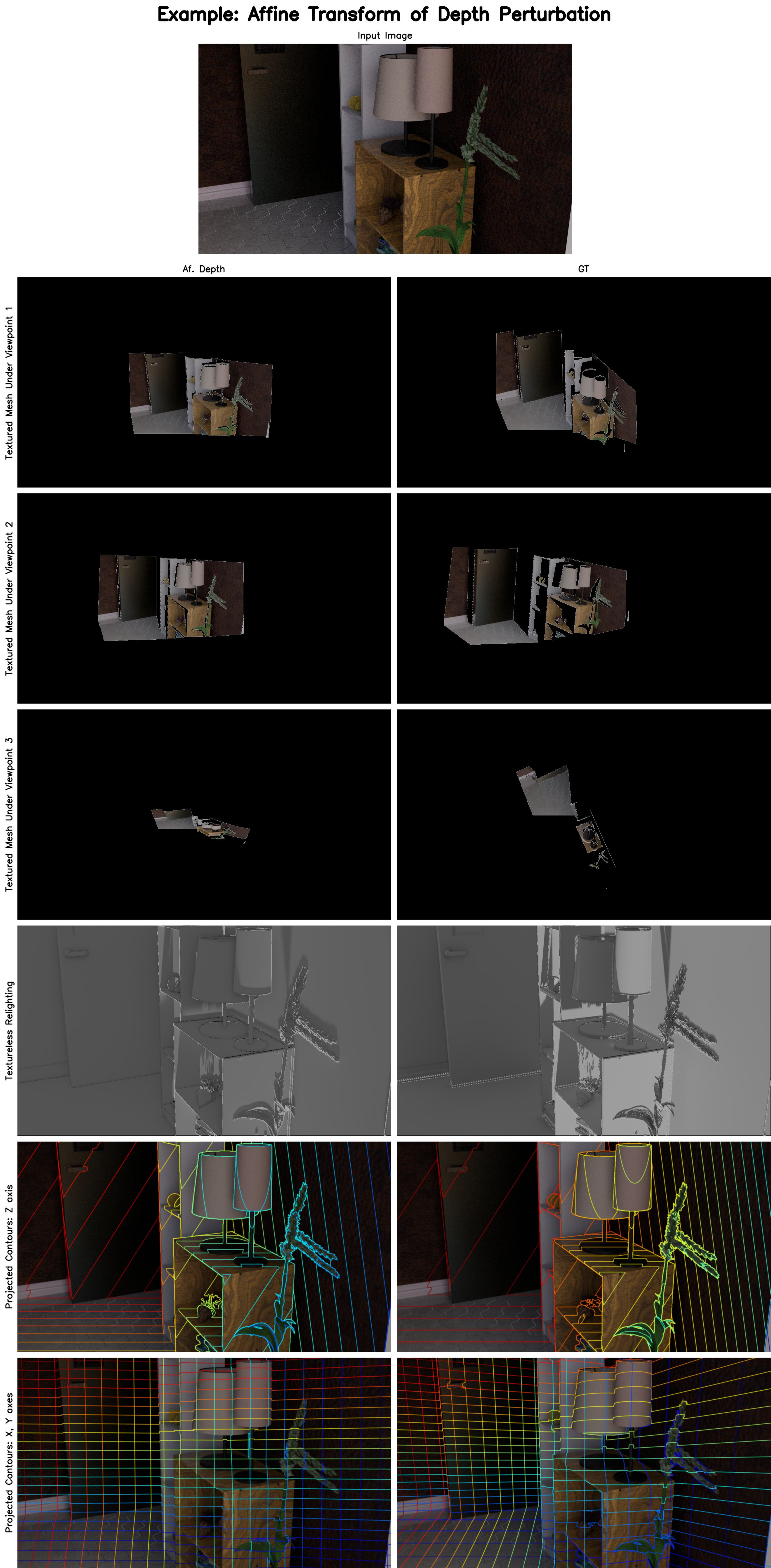}
    \caption{Example of Affine Transform of Depth Perturbation. In visualization of the textured mesh, images on the same row are captured under the same camera pose and using ground truth camera intrinsics. Depth is scaled by 0.2x before being translated to retain the same median depth, so the perturbed geometry is flattened}
    \label{fig:real_perturbation_example_af_depth}
\end{figure}

\begin{figure}[h]
    \centering
    \includegraphics[width=\linewidth]{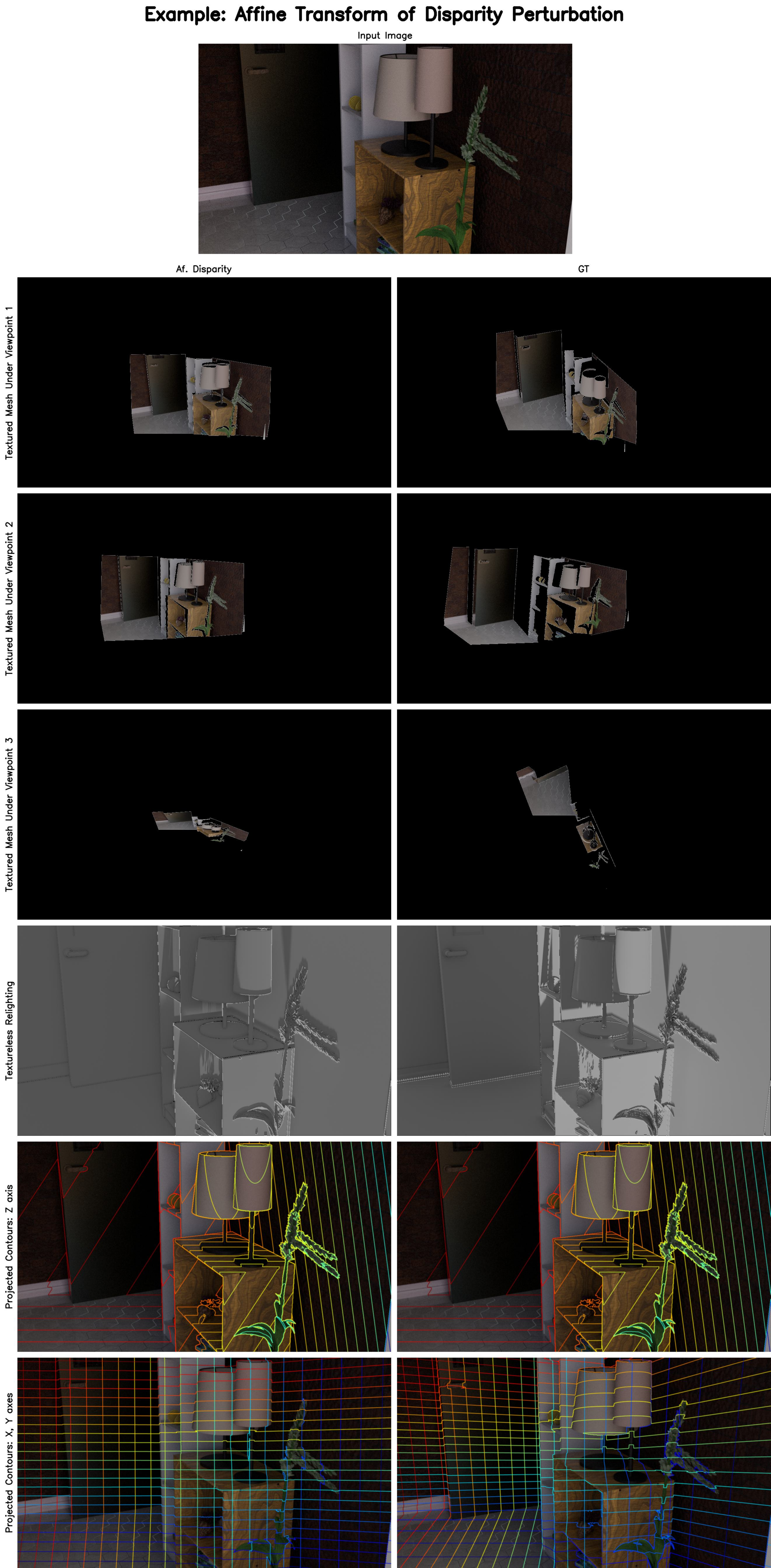}
    \caption{Example of Affine Transform of Disparity Perturbation. In visualization of the textured mesh, images on the same row are captured under the same camera pose and using ground truth camera intrinsics. Disparity is scaled by 0.2x before being translated to retain the same median disparity, so the perturbed geometry is flattened.}
    \label{fig:real_perturbation_example_af_disp}
\end{figure}

\begin{figure}[h]
    \centering
    \includegraphics[width=\linewidth]{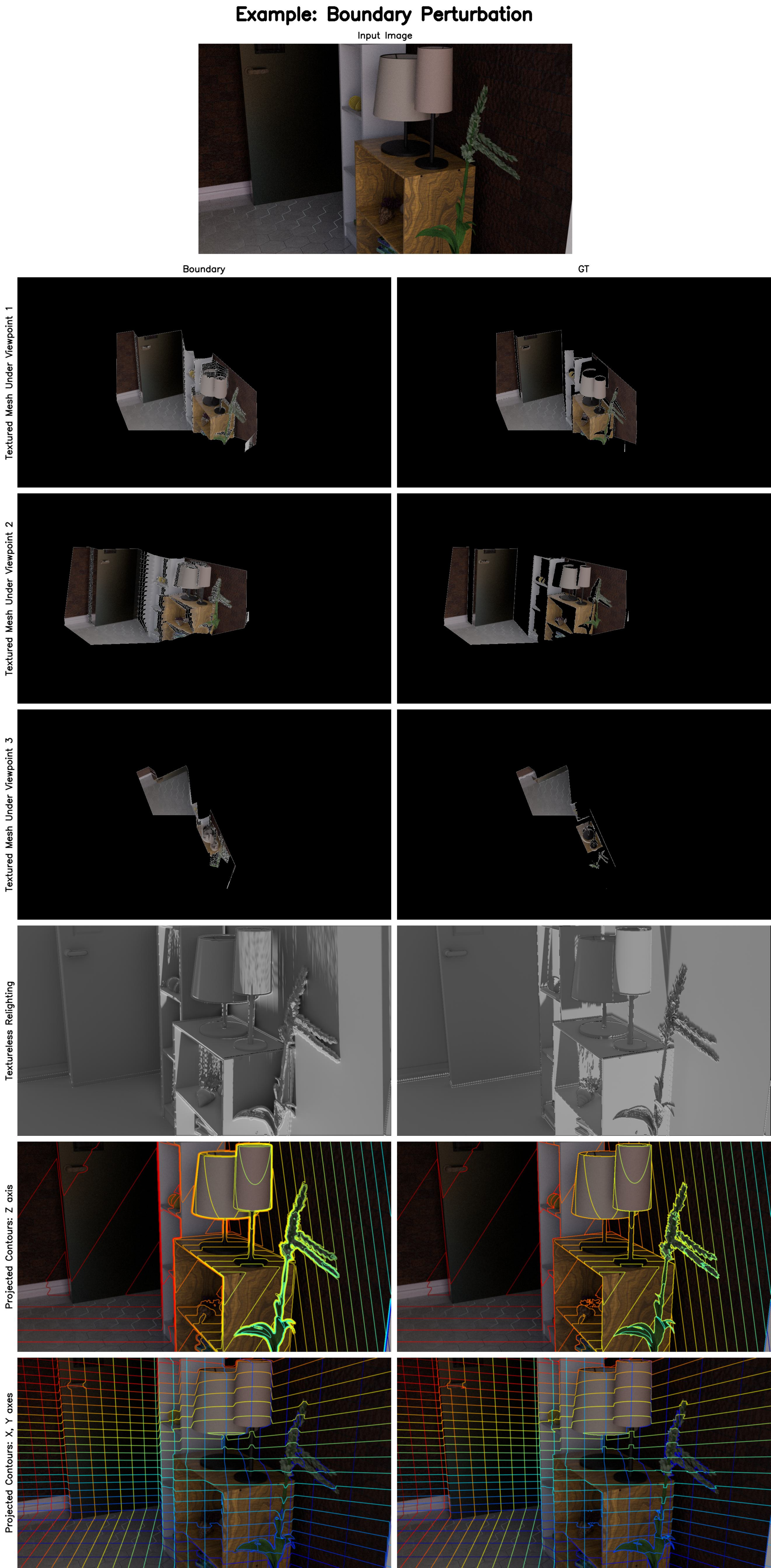}
    \caption{Example of Boundary Perturbation. In visualization of the textured mesh, images on the same row are captured under the same camera pose and using ground truth camera intrinsics. Depth values are blurred along boundaries, which can be easily observed in the textured mesh under different viewpoints and the thicker contour lines in Projected Contours}
    \label{fig:real_perturbation_example_bnd}
\end{figure}

Fig.~\ref{fig:gallary_of_dataset} shows images of 20 scenes randomly selected from the dataset. Fig.~\ref{fig:real_perturbation_example_surf_ori},\ref{fig:real_perturbation_example_cam_intr},\ref{fig:real_perturbation_example_rel_sc},\ref{fig:real_perturbation_example_curv_high_freq},\ref{fig:real_perturbation_example_curv_low_freq},\ref{fig:real_perturbation_example_af_depth},\ref{fig:real_perturbation_example_af_disp},\ref{fig:real_perturbation_example_bnd} show examples of every perturbation type. To make it easier to see how the geometry is perturbed, we present visualization of ground truth depth next to visualization of perturbed depth.

\section{Performance of MDE Methods under SAWA-H}
\label{sec:appx_eval_mde_methods}
\begin{figure*}[h]
    \centering
    \includegraphics[width=\linewidth]{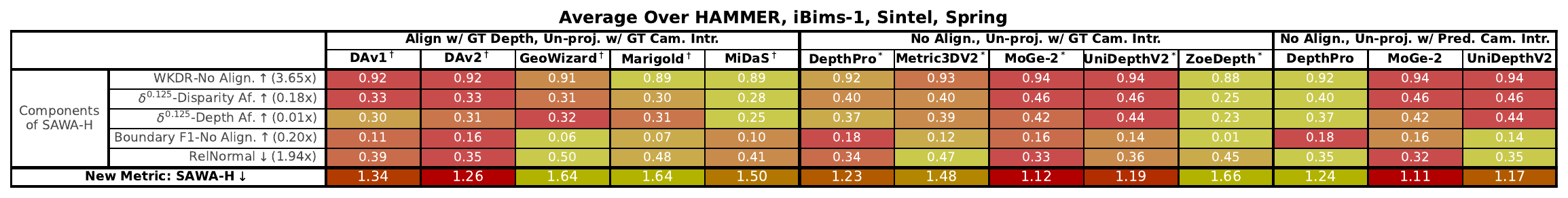}
    \caption{Performance of state-of-the-art monocular depth estimation methods under SAWA-H, averaged on datasets with accurate depth values near boundaries.}
    \label{fig:eval_mde_methods_with_sawah_bnd}
\end{figure*}

Fig.~\ref{fig:eval_mde_methods_with_sawah_detail} displays performance of state-of-the-art monocular depth estimation methods evaluated by SAWA-H on each dataset.
\begin{figure*}[h]
    \centering
    \includegraphics[width=\linewidth]{figures/eval_model_res-per_dataset.pdf}
    \caption{Performance of state-of-the-art monocular depth estimation methods under SAWA-H on each dataset.}
    \label{fig:eval_mde_methods_with_sawah_detail}
\end{figure*}
Fig.~\ref{fig:eval_mde_methods_with_sawah_bnd} displays performance of state-of-the-art monocular depth estimation methods evaluated by SAWA-H averaged on datasets with accurate depth values near boundaries.

\end{document}